\documentclass{article}

\usepackage{microtype}
\usepackage{graphicx}
\usepackage{subcaption}
\usepackage{booktabs} 
\usepackage{makecell}

\usepackage{multirow}
\usepackage{colortbl}
\usepackage{xcolor}
\definecolor{Gray}{gray}{0.9}

\usepackage{hyperref}

\newcommand{\inc}[1]{\textsubscript{$\uparrow$#1}}
\newcommand{\dec}[1]{\textsubscript{$\downarrow$#1}}

\usepackage[preprint]{icml2026}

\usepackage{amsmath}
\usepackage{amssymb}
\usepackage{mathtools}
\usepackage{amsthm}
\usepackage[normalem]{ulem}

\usepackage[capitalize,noabbrev]{cleveref}

\theoremstyle{plain}

\theoremstyle{definition}

\theoremstyle{remark}

\usepackage[textsize=tiny]{todonotes}

\icmltitlerunning{OpenMAG: A Comprehensive Benchmark for Multimodal-Attributed Graph}

\begin{document}

\twocolumn[
  \icmltitle{OpenMAG: A Comprehensive Benchmark for Multimodal-Attributed Graph}



  \icmlsetsymbol{equal}{*}

  \begin{icmlauthorlist}
    \icmlauthor{Chenxi Wan}{equal,yyy}
    \icmlauthor{Xunkai Li}{equal,yyy}
    \icmlauthor{Yilong Zuo}{yyy}
    \icmlauthor{Haokun Deng}{yyy}
    \icmlauthor{Sihan Li}{yyy}
    \icmlauthor{Bowen Fan}{yyy}
    \icmlauthor{Hongchao Qin}{yyy}
    \icmlauthor{Ronghua Li}{yyy}
    \icmlauthor{Guoren Wang}{yyy}
  \end{icmlauthorlist}

  \icmlaffiliation{yyy}{Department of Computer Science, Beijing Institute of Technology, Beijing, China}
  \icmlcorrespondingauthor{Rong-Hua Li}{lironghuabit@126.com}

  \icmlkeywords{Machine Learning, ICML}

  \vskip 0.3in
]



\printAffiliationsAndNotice{}  

\begin{abstract}
  Multimodal-Attributed Graph (MAG) learning has achieved remarkable success in modeling complex real-world systems by integrating graph topology with rich attributes from multiple modalities. With the rapid proliferation of novel MAG models capable of handling intricate cross-modal semantics and structural dependencies, establishing a rigorous and unified evaluation standard has become imperative. Although existing benchmarks have facilitated initial progress, they exhibit critical limitations in \textit{domain coverage}, \textit{encoder flexibility}, \textit{model diversity}, and \textit{task scope}, presenting significant challenges to fair evaluation. To bridge this gap, we present OpenMAG, a comprehensive benchmark that integrates 19 datasets across 6 domains and incorporates 16 encoders to support both static and trainable feature encoding. OpenMAG further implements a standardized library of 24 state-of-the-art models and supports 8 downstream tasks, enabling fair comparisons within a unified framework. Through systematic assessment of necessity, data quality, effectiveness, robustness, and efficiency, we derive 14 fundamental insights into MAG learning to guide future advancements. Our code is available at \url{https://anonymous.4open.science/r/OpenMAG-F703/}.
\end{abstract}

\section{INTRODUCTION}

With the rapid advancements of multimodal learning, Multimodal-Attributed Graphs (MAGs) have emerged as a critical framework for modeling complex real-world systems by unifying topological dependencies with rich semantic content from diverse modalities (e.g., texts, images). Specifically, MAGs not only enhance graph-based tasks by incorporating rich multimodal attributes, but also facilitate multimodal alignment and generation tasks by leveraging topological structures. This versatility demonstrates immense potential in high-value applications, ranging from recommendation systems~\cite{MMGCN,LGMRec} to social networks~\cite{DBLP:conf/www/Fan0LHZTY19,DBLP:conf/aaai/CaiWLZ024}.

Although MAG learning holds great promise, it faces significant challenges in handling the complex graph topology, aligning cross-modal semantics, and adapting to diverse downstream applications~\cite{DBLP:journals/corr/abs-2402-05322}. Consequently, numerous innovative models and methodologies have recently emerged to tackle these complexities~\cite{GraphGPT-O,MLaGA,DMGC}. This surge in innovation calls for a comprehensive benchmark to verify performance and guide future research.
While Existing benchmarks such as MM-GRAPH~\cite{mmgraph} and MAGB~\cite{MAGB} have laid the foundation, they fail to keep up with these evolving demands and exhibit critical limitations in four key dimensions: (1) \textbf{Datasets}: Limited to specific application domains. (2) \textbf{Feature Encoders}: Relying on frozen encoders, ignoring trainable and fine-tuning encoders. (3) \textbf{Model Backbones}: Lacking a unified evaluation to assess advanced multimodal alignment and fusion mechanisms~\cite{MMGCN,MGAT,UniGraph2}, failing to capture the interplay between cross-modal interactions and graph topology. (4) \textbf{Downstream Tasks}: Restricted to supervised graph-based tasks, overlooking unsupervised scenarios and the potential of graph structures in multimodal understanding and generative tasks.
We also note that GraphMLLM~\cite{GraphMLLM} introduces a benchmark for MAG learning; however, we exclude it from this comparison as it specifically centers on the evaluation of MLLM utilization paradigms.
These limitations not only restrict the assessment of model versatility, but also hinder the exploration of advanced paradigms like generative graph learning. Thus, there is an urgent need for a unified and comprehensive benchmark to standardize evaluation and foster future innovations in the MAG community.

\begin{table*}[t]
\centering
\caption{Comparison of different MAG benchmarks. \textbf{Graph-B.} and \textbf{Modal-B.} denote graph-based and modality-based tasks. \textbf{Effe.}, \textbf{Effi.}, \textbf{Pref.}, \textbf{Neces.}, \textbf{Rob.}, and \textbf{Qual.} denote Effectiveness, Efficiency, Modality Preference, Necessity, Robustness, and Data Quality.}
\label{benchmark_comparison}
\resizebox{\textwidth}{!}{%
\begin{tabular}{c|c|c|c|c|c}
\toprule
\textbf{MAG Benchmarks} & \textbf{Datasets} & \textbf{Encoders} & \textbf{Model Backbones} & \textbf{Downstream Tasks} & \textbf{Evaluations} \\
\midrule
MAGB[KDD'25] & 5 (2 Domains) & Frozen Only & 4 (Unimodal GNNs) & Supervised Graph-B. & Effe.+Effi.+Pref. \\
MM-GRAPH[CVPR'25] & 7 (3 Domains) & Frozen Only & 7 (Uni+Multimodal) & Supervised Graph-B. & Effe.+Qual.+Neces. \\
\midrule
OpenMAG (Ours) & 19 (6 Domains) & Frozen+Trainable & \makecell[c]{24 (Graph+Multimodal \\ +LLM-enhanced)} & \makecell[c]{(Un)Supervised \\Graph-B. +Modal-B.} & \makecell[c]{Neces.+Qual.+Effe. \\ +Rob.+Effi.} \\
\bottomrule
\end{tabular}%
}
\end{table*}

To bridge these gaps, we present OpenMAG, a unified benchmark integrating 19 datasets across 6 domains, 16 modality encoders ranging from frozen extractors to trainable backbones, and 24 graph learning models to support 3 graph-based and 5 modality-based tasks. Based on this design, we establish a systematic evaluation framework for MAG learning from five critical perspectives. 
For \textbf{Necessity}, we validate the fundamental premise of MAG learning by confirming the mutual indispensability of multimodal semantics and topological structure. 
For \textbf{Data Quality}, we investigate how modality encoding strategies determine the semantic foundation quality required for effective downstream reasoning.
For \textbf{Effectiveness}, we evaluate model capabilities across a broad spectrum of tasks ranging from discriminative reasoning to generative content creation to establish a comprehensive standard for model superiority. 
For \textbf{Robustness}, we assess model resilience against real-world imperfections to gauge reliability in non-ideal deployment environments. 
For \textbf{Efficiency}, we analyze the critical trade-off between computational resource consumption and performance gains to guide the development of scalable algorithms. 
To further emphasize the distinct advantages of our proposed OpenMAG compared to current benchmarks, we provide a detailed comparison in Table~\ref{benchmark_comparison}.

\textbf{Our Contributions}. 
(1) \underline{\textit{Comprehensive Benchmark}}. We propose OpenMAG, which integrates 19 datasets in 6 different domains. For multimodal encoding methods, we support 16 encoders ranging from frozen extractors to trainable backbones. By standardizing 24 MAG learning models and 8 downstream tasks, OpenMAG serves as a comprehensive framework for the standardized evaluation of MAG learning methods.
(2) \underline{\textit{Valuable Insights}}. We conduct systematic experiments from 5 valuable perspectives and summarize 14 key conclusions. We distill these findings into actionable insights that offer a clear roadmap for future research.
(3) \underline{\textit{Open-sourced Benchmark Library}}. We release OpenMAG as an open-source and user-friendly library, enabling researchers to easily evaluate custom methods or datasets. In addition, we provide comprehensive documentation and modular interfaces to foster collaborative innovation and facilitate the growth of the MAG community. The code and resources are available at \url{https://anonymous.4open.science/r/OpenMAG-F703/}.

\begin{figure*}[t!]
    \centering
    \includegraphics[width=1.0\linewidth]{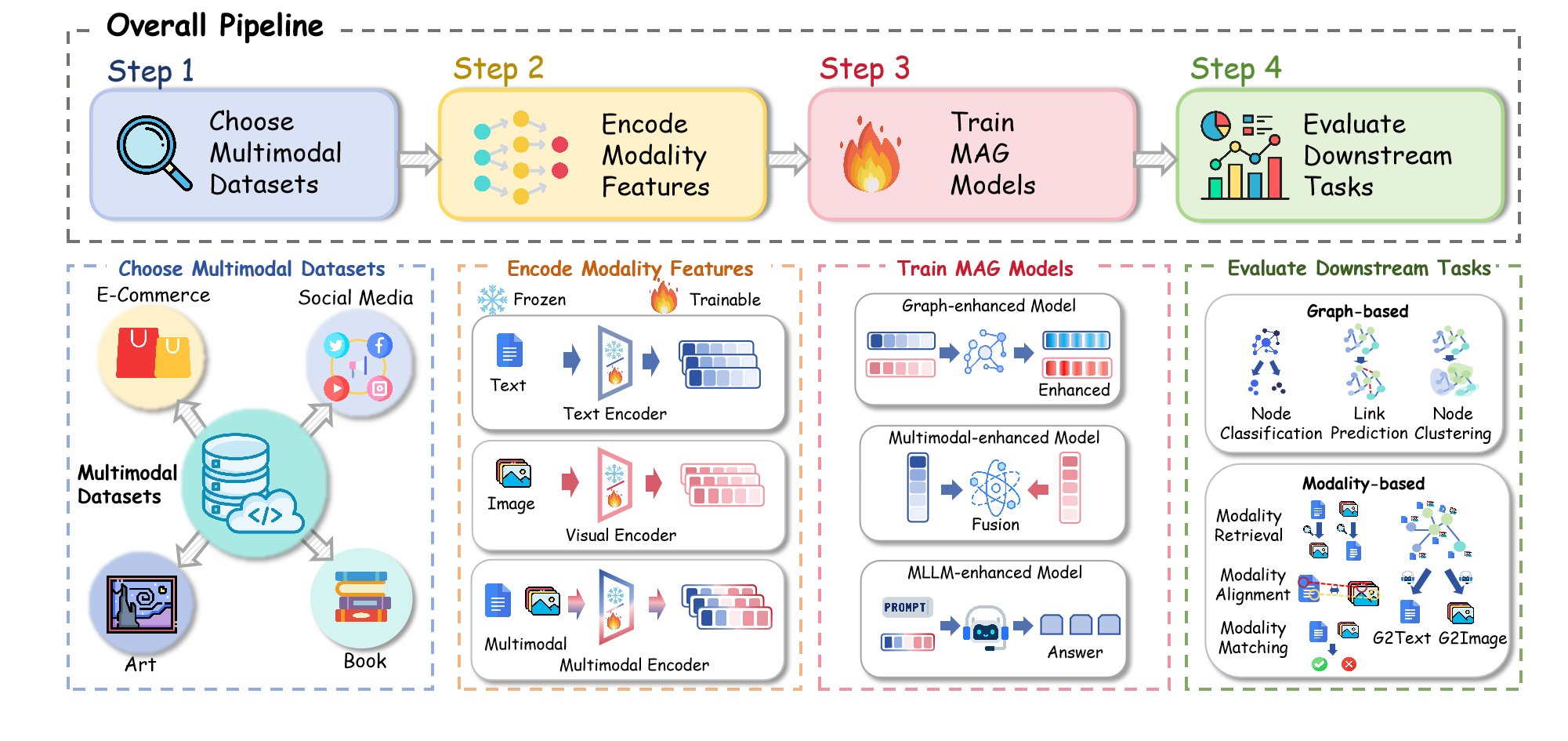}
    \caption{The overall framework of OpenMAG.}
    \label{fig:framework}
\end{figure*}

\section{Preliminaries}
\label{sec:preliminaries}

\subsection{Notations and Learning Paradigms}
\label{sec:notations_paradigms}

Formally, a Multimodal-Attributed Graph (MAG) is defined as $\mathcal{G}=(\mathcal{V}, \mathcal{E}, \mathcal{X}_{\mathcal{M}})$, where $\mathcal{V}$ is the node set, $\mathcal{E}$ is the edge set, and $\mathcal{X}_{\mathcal{M}}$ represents raw attributes from modalities $\mathcal{M}=\{T, V\}$ (Text, Visual). Unlike traditional multimodal learning, MAG learning leverages topology to enhance representation via neighbor interaction. The pipeline proceeds in four stages: 

\textbf{Step 1: Multimodal Encoding.} Raw attributes $x_v^m \in \mathcal{X}_{\mathcal{M}}$ are projected into initial embeddings $\mathbf{h}_v^{m, (0)}$ via modality-specific encoders $f_{\theta}^m$: $\mathbf{h}_v^{m, (0)} = f_{\theta}^m(x_v^m)$. 

\textbf{Step 2: Multimodal Fusion.} Heterogeneous features are integrated into a unified node representation $\mathbf{z}_v$ using a fusion operator $\Phi$: $\mathbf{z}_v = \Phi(\{\mathbf{h}_v^{m, (0)}\}_{m \in \mathcal{M}})$. 

\textbf{Step 3: Multimodal Interaction.} Node embeddings are updated by aggregating topological neighbors $\mathcal{N}(v)$ through $L$ layers of structure-aware propagation: $\mathbf{h}_v^{(l)} = \text{AGG}(\mathbf{h}_v^{(l-1)}, \{\mathbf{h}_u^{(l-1)}\}_{u \in \mathcal{N}(v)})$. 

\textbf{Step 4: Multimodal Adaptation.} The final representations $\mathbf{h}_v^{(L)}$ are mapped to the task-specific output space $\hat{Y}$ via a projection head $\psi$: $\hat{y}_v = \psi(\mathbf{h}_v^{(L)})$.

\subsection{Problem Formulation}
\label{sec:problem_formulation}

OpenMAG comprehensively evaluates models across a diverse spectrum of 8 
downstream tasks, categorized into two groups. \textbf{Graph-Based Tasks} 
include Node Classification, Link Prediction, and Node Clustering. These 
tasks focus on learning discriminative representations from graph topology 
and attributes to distinguish node properties or community structures. 
\textbf{Modality-Based Tasks} comprise Modality Matching, Modality 
Retrieval, Modality Alignment, Graph-to-Text Generation (G2Text), and 
Graph-to-Image Generation (G2Image). These tasks focus on verifying 
cross-modal semantic consistency and generating high-fidelity content 
conditioned on graph contexts. Detailed definitions and metrics for each 
task are provided in Appendix~\ref{app:downstream_tasks}.

\section{BENCHMARK DESIGN}
\label{sec:benchmark_design}

In this section, we present a comprehensive and structural overview of the OpenMAG design. As illustrated in Figure~\ref{fig:framework}, OpenMAG unifies the entire research lifecycle into a standardized pipeline. First, we detail the dataset repository spanning diverse real-world domains (Sec.~\ref{sec:dataset}). Next, we introduce the versatile encoder module for semantic representation extraction (Sec.~\ref{sec:encoder}), followed by a standardized library of representative MAG models (Sec.~\ref{sec:model}). Lastly, we outline 5 dimensional evaluation protocol (Sec.~\ref{sec:evaluation}) that structures the rigorous evaluation process to ensure fair comparisons.

\subsection{Dataset Overview For OpenMAG}
\label{sec:dataset}
To address the lack of domain diversity in existing benchmarks, we choose a rich repository of datasets spanning e-commerce products~\cite{Amazon2018,Amazonreview1,Amazonreview2}, social media~\cite{RedditS}, art networks~\cite{semart}, video recommendation~\cite{ninerec}, book recommendation~\cite{books-nc1,books-nc2} and image networks~\cite{flickr30k}. OpenMAG employs domain-specific strategies to construct graph topologies that align with the distinct characteristics of each scenario. For social media, we define relationships through co-viewing patterns; and for e-commerce, we establish edges based on co-purchase records. By integrating such diverse scenarios, OpenMAG ensures that model performance is evaluated against the complexity of real-world distributions rather than overfitting to specific datasets. Detailed statistics and domain descriptions are provided in Appendix~\ref{app:datasets}.

\subsection{Encoder Module For OpenMAG}
\label{sec:encoder}

To extract high-quality semantic representations from raw and unstructured multimodal attributes, OpenMAG provides a versatile encoder module that supports both frozen feature encoding and adaptable end-to-end fine-tuning strategies.

\textbf{Frozen Encoders.} For text attributes, the library covers discriminative models like Sentence BERT~\cite{sbert}, T5~\cite{t5} and generative LLMs like Llama3.2~\cite{llama3.2}. For visual attributes, we incorporate diverse vision models such as ViT~\cite{vit,imagenet}, DINOv2~\cite{dinov2} and Swinv2~\cite{swin}. Furthermore, to leverage native cross-modal alignment, we include VLMs such as CLIP~\cite{clip} and Qwen-VL~\cite{qwenvl}. Detailed specifications of all integrated backbones are provided in Appendix~\ref{app:feature-encoders}.

\textbf{Trainable Encoders.} Static embeddings often yield general-purpose representations that lack specificity for diverse downstream graph tasks. To address this, OpenMAG explicitly fine-tunes representative backbones, specifically BERT and ViT, to support end-to-end optimization. This paradigm enables gradient propagation from downstream objectives back to encoders, allowing the semantic space to dynamically adapt to the target distribution for more rigorous task-specific assessment.

\subsection{Model Library For OpenMAG}
\label{sec:model}
OpenMAG integrates a standardized library of state-of-the-art MAG learning models. Excluding traditional unimodal GNNs~\cite{GCN,GAT,GraphSAGE} which serve as baselines, we categorize the advanced MAG-specific models into a unified taxonomy comprising three distinct paradigms: \textbf{Graph-enhanced Models} that prioritize topological representation learning by employing refined message-passing architectures or structural reconstruction to capture intricate dependencies~\cite{DMGC,DGF}; \textbf{Multimodal-enhanced Models} that incorporate specialized mechanisms to capture cross-modal interactions~\cite{MMGCN,MGAT}; and \textbf{MLLM-enhanced} Models that leverage LLMs to align graph topology with semantic spaces by treating structural information as auxiliary tokens~\cite{UniGraph2,GraphGPT-O,MLaGA}. By implementing these diverse paradigms within a single framework, OpenMAG enables a systematic analysis of how distinct paradigms perform in diverse downstream scenarios. Detailed information about these models can be found in Appendix~\ref{app:model_framework}.

\subsection{Task Evaluation For OpenMAG}
\label{sec:evaluation}

To comprehensively evaluate the proposed OpenMAG pipeline, we introduce 5 evaluation protocols encompassing \textbf{Necessity}, \textbf{Data Quality}, \textbf{Effectiveness}, \textbf{Robustness}, and \textbf{Efficiency}. Each dimension incorporates specialized assessment strategies tailored to OpenMAG’s objective of establishing a rigorous and universal evaluation standard.

\textbf{Necessity.} We verify the mutual indispensability of multimodal semantics and graph topology. We quantify the gains from integrating multimodal features over unimodal baselines. Besides, we assess how the incorporation of graph structure enhances consistency in multimodal generation tasks. This validates the synergy between semantic content and structural topology is essential in MAG learning.

\textbf{Data Quality.} We evaluate the quality of embeddings derived from diverse encoder architectures, specifically analyzing the alignment quality of unified multimodal encoders versus independent text and visual encoders. We assess this alignment through modality retrieval tasks using MRR and Hits@K metrics. Furthermore, we conduct a comparative analysis between end-to-end fine-tuning and frozen parameter paradigms on node classification tasks using Accuracy and F1-score. This dimension identifies the optimal configuration to generate high-quality embeddings.

\textbf{Effectiveness.} We conduct a systematic evaluation of diverse models across graph-based tasks and modality-based tasks to assess both reasoning and generative capabilities. Furthermore, we investigate the impact of advanced fine-tuning strategies on enhancing the generative quality of LLMs. This dimension provides a comprehensive assessment of model capabilities, ensuring a rigorous validation of their effectiveness across diverse learning objectives.

\textbf{Robustness.} We test model stability against real-world imperfections by injecting modal noise and simulating structural sparsity. By observing performance degradation under these perturbations, we evaluate the resilience of different architectures. This dimension highlights models that maintain reliability in noisy or incomplete data environments.

\textbf{Efficiency.} We assess the computational cost regarding both time and space consumption. For time efficiency, we derive the theoretical time complexity and record the number of epochs required for convergence. For space efficiency, we measure the peak GPU memory usage to evaluate the memory overhead. This dimension identifies the optimal trade-off between resource consumption and model performance for practical deployment.

\section{EXPERIMENTS AND ANALYSIS}
\label{sec:exp}

In this section, we conduct extensive experiments using OpenMAG to provide a comprehensive evaluation of MAG learning methods. Our analysis is driven by ten questions centered on five critical dimensions: (1) \textbf{For Necessity}, \textbf{Q1}: Does the integration of multimodal information improve performance compared to unimodal baselines? \textbf{Q2}: Can graph structure effectively enhance the consistency of multimodal generative tasks? (2) \textbf{For Data Quality}, \textbf{Q3}: Does end-to-end fine-tuning of encoders lead to significant performance gains compared to using frozen parameters? \textbf{Q4}: How do different feature encoding strategies impact the quality of multimodal embeddings? (3) \textbf{For Effectiveness}, \textbf{Q5}: How do different MAG learning models perform in different downstream scenarios? \textbf{Q6}: Do advanced parameter-efficient fine-tuning strategies significantly boost the generative capabilities of downstream models? (4) \textbf{For Robustness}, \textbf{Q7}: Which model paradigm demonstrates superior resilience against modality and label noise? \textbf{Q8}: How does model performance degrade under structural perturbation scenarios (5) \textbf{For Efficiency}, \textbf{Q9}: What are the comparative theoretical time and space complexities of different models? \textbf{Q10}: What are the actual empirical computational costs on downstream tasks? More detailed experimental setups are provided in Appendix~\ref{app:env} and~\ref{app:hyperparameters}.


\subsection{Validation of Structure-Modality Necessity}

This section investigates the mutual reinforcement between modality and topology: whether multimodal integration enhances graph learning (Q1), and conversely, whether topological structure improves multimodal generation (Q2). Detailed experimental setups are provided in Appendix~\ref{app:hyperparameters}.

\begin{figure}[t]
    \centering
    \includegraphics[width=0.95\linewidth]{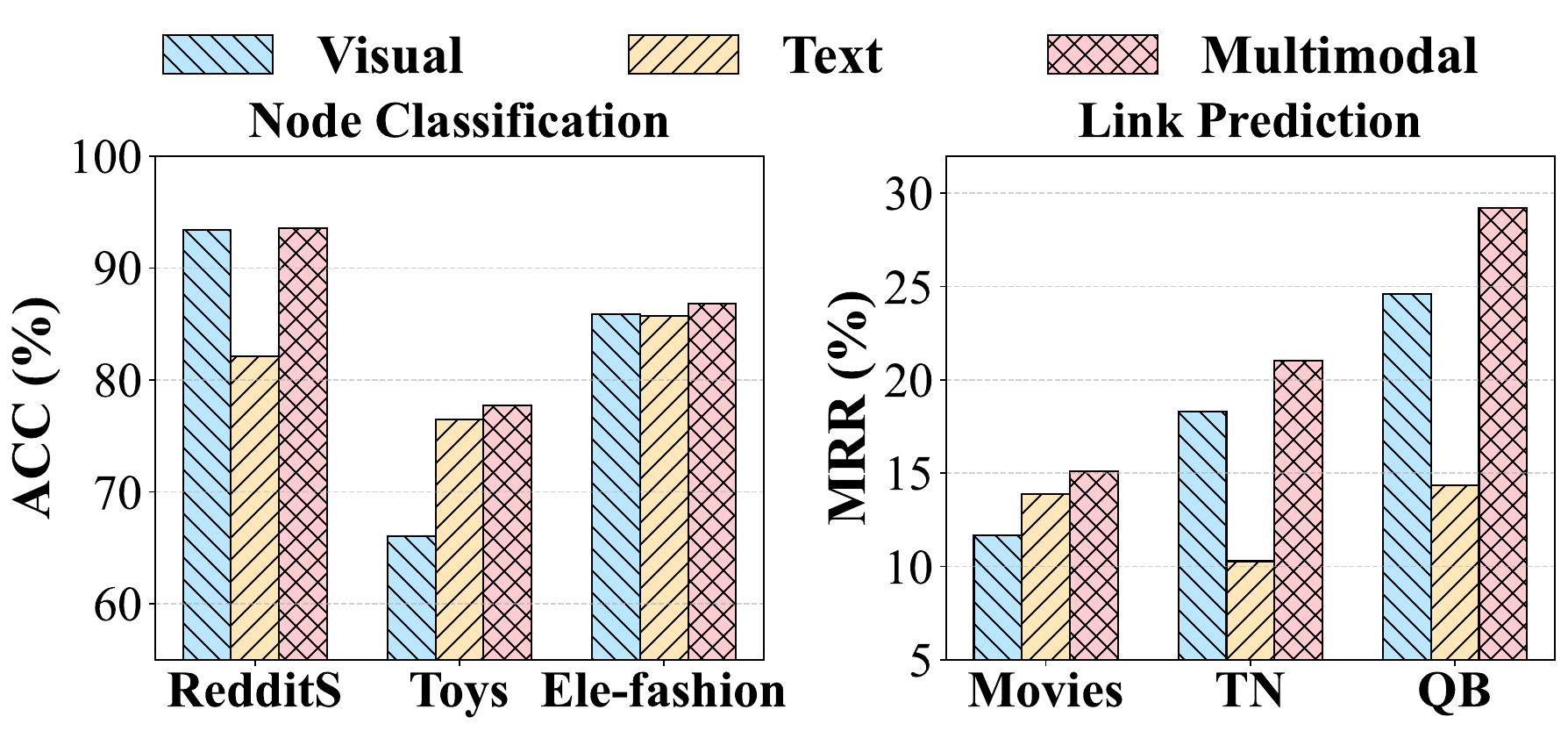}
    \caption{Performance comparison on unimodal and multimodal settings.}
    \label{fig:Q1}
\end{figure}

To address \textbf{Q1}, we comprehensively analyze unimodal and multimodal approaches to evaluate the necessity of integrating diverse modalities. As illustrated in Figure~\ref{fig:Q1}, the multimodal setting consistently outperforms unimodal baselines across all datasets, demonstrating that fusion of distinct modalities enriches node representations. Notably, we observe domain-specific modality preference: textual features excel in e-commerce datasets due to product descriptions, while visual features dominate in social scenarios by driving user interaction. This highlights the complementary nature of diverse modalities across application contexts.

\begin{table}[!ht]
\centering
\caption{Evaluation of Graph Structure Effectiveness on Multimodal Generative Task.}
\label{tab:Q2}
\resizebox{\columnwidth}{!}{
\begin{tabular}{c|cc|cc}
\toprule

\multirow{2}{*}{\textbf{Neighbors}} 
& \multicolumn{2}{c|}{\textbf{SemArt}} 
& \multicolumn{2}{c}{\textbf{Toys}} \\

\cmidrule{2-5}

& \textbf{CLIP-Score} & \textbf{DINO-Score}
& \textbf{CLIP-Score} & \textbf{DINO-Score} \\

\midrule

$0$ 
& $68.58$ & $50.86$ 
& $57.83$ & $28.13$ \\
\midrule

$2$ 
& $68.77\inc{0.19}$ & $53.20\inc{2.34}$ 
& $58.47\inc{0.64}$ & $29.16\inc{1.03}$ \\

$4$ 
& $70.21\inc{1.63}$ & $55.29\inc{4.43}$ 
& $59.43\inc{1.60}$ & $29.49\inc{1.36}$ \\
$6$ 
& $67.27\dec{1.31}$ & $51.41\inc{0.55}$ 
& $60.19\inc{2.36}$ & $30.83\inc{2.70}$ \\

$8$ 
& $67.10\dec{1.48}$ & $50.57\dec{0.29}$ 
& $59.06\inc{1.23}$ & $29.67\inc{1.54}$ \\

\bottomrule
\end{tabular}
}
\end{table}

To address \textbf{Q2}, we utilize the G2Image task on SemArt and Toys datasets to investigate the impact of graph structure on multimodal generation consistency. We explicitly control the number of aggregated neighbors to compare the setting without graph structure against settings with graph structure. Specifically, the setting with zero neighbors represents a structure-agnostic baseline where pre-trained embeddings are directly projected via a linear layer without topological interaction, whereas settings with non-zero neighbors employ a Graph-QFormer~\cite{instructg2i} to aggregate neighborhood information. As presented in Table~\ref{tab:Q2}, incorporating appropriate structural context significantly boosts performance compared to the baseline, acting as a semantic anchor to align generated content. However, performance degrades when neighbor aggregation exceeds an optimal threshold. This suggests that while moderate structural guidance enhances consistency, over-dense neighborhoods introduce semantic noise that dilutes distinct node features and compromises generation fidelity.

Based on the observations above, we summarize four key insights: \textbf{C1}: \textit{The integration of multimodal information is essential for enhancing the overall performance of multimodal attributed graph learning.}~\cite{mmgraph} \textbf{C2}: \textit{Different domains exhibit modality preferences, where textual features excel in e-commerce and visual features excel in social scenarios.}~\cite{MMGCN} \textbf{C3}: \textit{Graph structures enrich node semantics by aggregating neighbor information in MAG learning.}~\cite{GCN} \textbf{C4}: \textit{Excessive neighbors may introduce semantic noise that compromises the generation quality.}~\cite{instructg2i}

\begin{figure*}[t]
    \centering
    \begin{minipage}{0.48\textwidth}
        \centering
        \includegraphics[width=\linewidth]{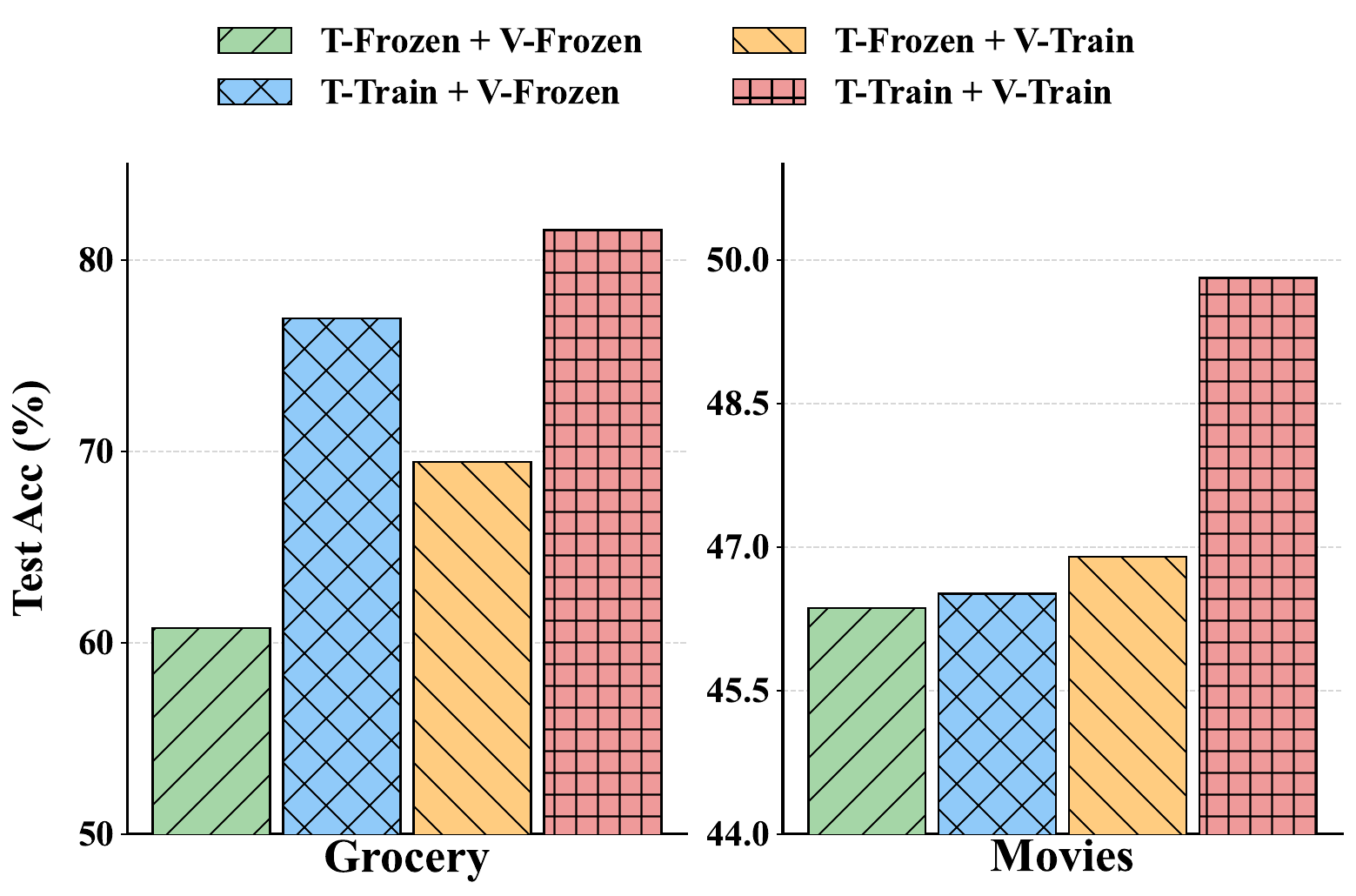}
        \vspace{-0.2cm}
        \caption{Impact of fine-tuning strategies.}
        \label{fig:q3_results}
    \end{minipage}
    \hfill
    \begin{minipage}{0.48\textwidth}
        \centering
        \includegraphics[width=\linewidth]{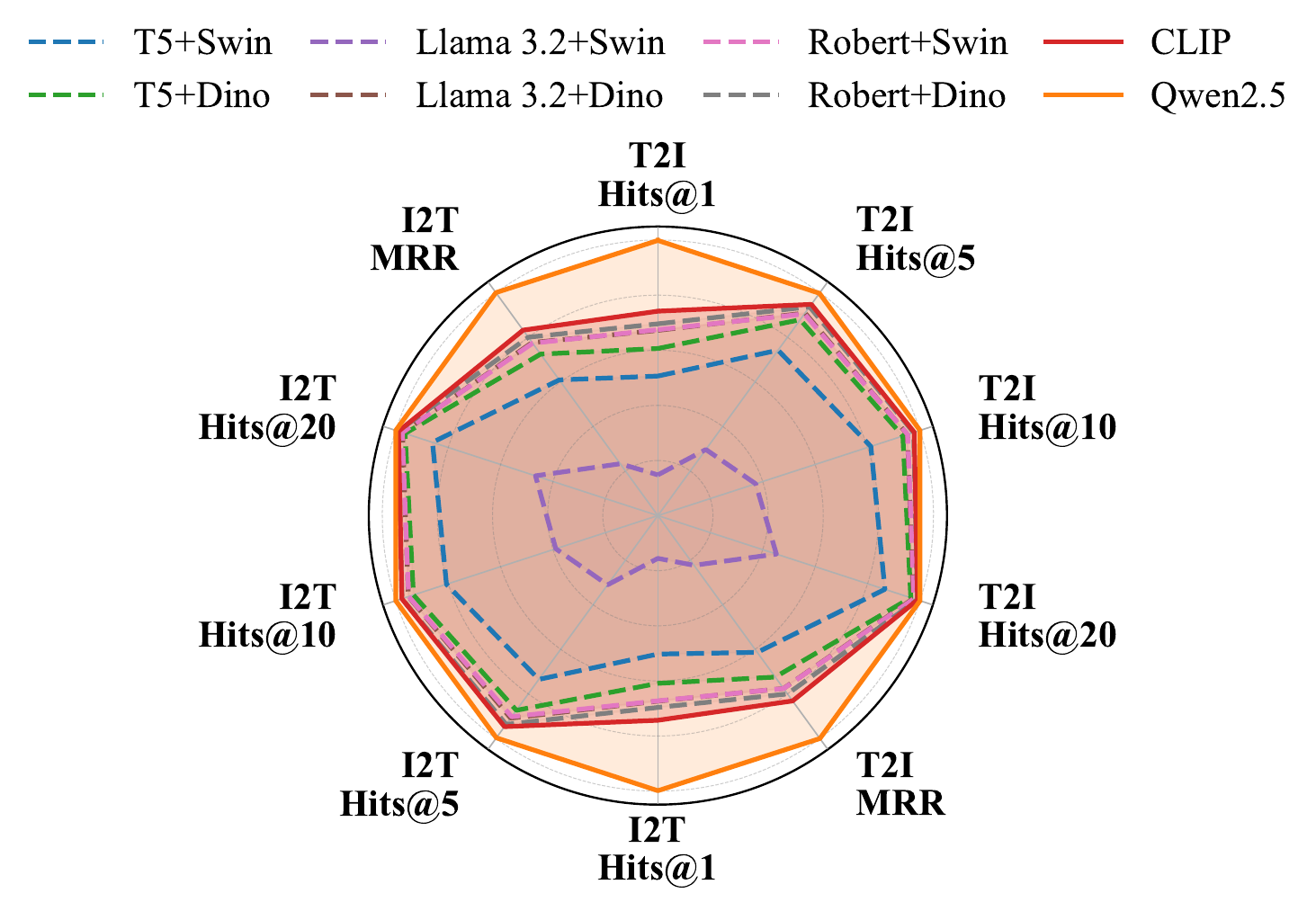}
        \vspace{-0.2cm}
        \caption{Comparison of encoding strategies.}
        \label{fig:q4_results}
    \end{minipage}
    \vspace{-0.3cm}
\end{figure*}

\begin{table*}[t!]
\centering
\caption{Model Performance in Graph-Based Downstream Tasks. The best result is $\textbf{bold}$. The second result is $\underline{underlined}$.}
\label{tab:Q5_1}
\resizebox{\textwidth}{!}{
\begin{tabular}{c|cccc|cccc|cccc}
\toprule

\multirow{3}{*}{\centering\textbf{Methods}}
& \multicolumn{4}{c|}{\textbf{Node Classification}}
& \multicolumn{4}{c|}{\textbf{Link Prediction}}
& \multicolumn{4}{c}{\textbf{Node Clustering}} \\

\cmidrule{2-13}

& \multicolumn{2}{c}{Movies} 
& \multicolumn{2}{c|}{Grocery}
& \multicolumn{2}{c}{DY} 
& \multicolumn{2}{c|}{Bili\_Dance}
& \multicolumn{2}{c}{Toys}
& \multicolumn{2}{c}{RedditS} \\

& Acc & F1-score
& Acc & F1-score
& MRR & Hits@3
& MRR & Hits@3
& NMI & ARI
& NMI & ARI \\

\midrule

MLP
& $51.22_{\scriptsize \pm 0.12}$ & $39.60_{\scriptsize \pm 0.92}$ & $82.60_{\scriptsize \pm 0.26}$ & $74.80_{\scriptsize \pm 0.36}$
& $62.57_{\scriptsize \pm 0.89}$ & $75.65_{\scriptsize \pm 1.23}$ & $28.90_{\scriptsize \pm 0.27}$ & $34.87_{\scriptsize \pm 0.41}$
& $45.47_{\scriptsize \pm 0.68}$ & $29.20_{\scriptsize \pm 1.05}$ & $77.78_{\scriptsize \pm 0.15}$ & $69.90_{\scriptsize \pm 2.27}$ \\

GCN
& $\underline{52.84}_{\scriptsize \pm 0.40}$ & $\underline{44.78}_{\scriptsize \pm 0.44}$ & $80.37_{\scriptsize \pm 0.45}$ & $71.89_{\scriptsize \pm 0.39}$
& $70.28_{\scriptsize \pm 0.15}$ & $84.21_{\scriptsize \pm 0.15}$ & $37.79_{\scriptsize \pm 0.29}$ & $47.67_{\scriptsize \pm 0.27}$
& $48.03_{\scriptsize \pm 0.71}$ & $32.31_{\scriptsize \pm 1.45}$ & $78.39_{\scriptsize \pm 0.52}$ & $71.21_{\scriptsize \pm 2.11}$ \\

GAT
& $51.38_{\scriptsize \pm 0.64}$ & $43.22_{\scriptsize \pm 0.31}$ & $80.26_{\scriptsize \pm 0.32}$ & $72.61_{\scriptsize \pm 0.32}$
& $70.67_{\scriptsize \pm 0.23}$ & $84.96_{\scriptsize \pm 0.28}$ & $36.85_{\scriptsize \pm 0.36}$ & $47.36_{\scriptsize \pm 0.81}$
& $48.66_{\scriptsize \pm 0.23}$ & $31.96_{\scriptsize \pm 1.19}$ & $78.40_{\scriptsize \pm 1.41}$ & $68.90_{\scriptsize \pm 4.38}$ \\

\midrule

DMGC
& $52.04_{\scriptsize \pm 0.85}$ & $41.99_{\scriptsize \pm 1.66}$ & $82.30_{\scriptsize \pm 0.51}$ & $70.90_{\scriptsize \pm 0.70}$
& $\underline{74.47}_{\scriptsize \pm 0.84}$ & $\underline{89.10}_{\scriptsize \pm 0.73}$ & $\underline{39.95}_{\scriptsize \pm 2.55}$ & $\underline{53.86}_{\scriptsize \pm 5.05}$
& $49.38_{\scriptsize \pm 2.28}$ & $32.51_{\scriptsize \pm 3.14}$ & $79.72_{\scriptsize \pm 1.34}$ & $66.08_{\scriptsize \pm 4.26}$ \\

DGF
& $\textbf{53.89}_{\scriptsize \pm 0.33}$ & $41.45_{\scriptsize \pm 0.48}$ & $\underline{82.92}_{\scriptsize \pm 0.16}$ & $71.83_{\scriptsize \pm 0.28}$
& $\textbf{77.28}_{\scriptsize \pm 0.17}$ & $\textbf{92.51}_{\scriptsize \pm 0.18}$ & $\textbf{42.55}_{\scriptsize \pm 0.12}$ & $\textbf{58.25}_{\scriptsize \pm 0.37}$
& $\textbf{51.29}_{\scriptsize \pm 0.62}$ & $\textbf{36.24}_{\scriptsize \pm 0.91}$ & $\textbf{84.89}_{\scriptsize \pm 0.36}$ & $\textbf{78.07}_{\scriptsize \pm 1.16}$ \\

\midrule

MMGCN
& $52.12_{\scriptsize \pm 0.98}$ & $41.85_{\scriptsize \pm 1.55}$ & $82.63_{\scriptsize \pm 0.24}$ & $75.03_{\scriptsize \pm 1.12}$
& $69.66_{\scriptsize \pm 0.42}$ & $84.17_{\scriptsize \pm 0.55}$ & $37.70_{\scriptsize \pm 0.68}$ & $49.81_{\scriptsize \pm 0.88}$
& $45.39_{\scriptsize \pm 0.93}$ & $27.56_{\scriptsize \pm 0.48}$ & $67.90_{\scriptsize \pm 2.62}$ & $50.42_{\scriptsize \pm 6.68}$ \\

MGAT
& $50.00_{\scriptsize \pm 0.15}$ & $37.31_{\scriptsize \pm 1.58}$ & $82.61_{\scriptsize \pm 0.24}$ & $\underline{75.10}_{\scriptsize \pm 0.33}$
& $70.37_{\scriptsize \pm 0.17}$ & $85.05_{\scriptsize \pm 0.30}$ & $36.74_{\scriptsize \pm 0.16}$ & $49.37_{\scriptsize \pm 0.69}$
& $46.72_{\scriptsize \pm 1.32}$ & $30.02_{\scriptsize \pm 1.63}$ & $73.36_{\scriptsize \pm 1.51}$ & $60.55_{\scriptsize \pm 4.17}$ \\

LGMRec
& $52.59_{\scriptsize \pm 0.25}$ & $\textbf{46.26}_{\scriptsize \pm 0.26}$
& $\textbf{84.27}_{\scriptsize \pm 0.04}$ & $\textbf{77.62}_{\scriptsize \pm 0.05}$
& $69.57_{\scriptsize \pm 0.04}$ & $85.08_{\scriptsize \pm 0.17}$
& $39.92_{\scriptsize \pm 0.40}$ & $51.85_{\scriptsize \pm 0.88}$
& $\underline{49.94}_{\scriptsize \pm 0.94}$ & $\underline{35.31}_{\scriptsize \pm 1.26}$
& $\underline{80.73}_{\scriptsize \pm 0.96}$ & $\underline{73.10}_{\scriptsize \pm 3.53}$ \\

\midrule

UniGraph2
& $46.78_{\scriptsize \pm 0.36}$ & $31.03_{\scriptsize \pm 1.57}$
& $75.86_{\scriptsize \pm 1.52}$ & $64.03_{\scriptsize \pm 3.40}$
& $65.78_{\scriptsize \pm 0.58}$ & $78.43_{\scriptsize \pm 0.64}$
& $31.58_{\scriptsize \pm 1.47}$ & $41.32_{\scriptsize \pm 2.58}$
& $10.34_{\scriptsize \pm 3.61}$ & $3.09_{\scriptsize \pm 1.51}$
& $31.85_{\scriptsize \pm 2.84}$ & $12.85_{\scriptsize \pm 2.29}$ \\

\bottomrule
\end{tabular}
}
\end{table*}

\begin{table*}[t!]
    \centering
    \begin{minipage}{0.52\linewidth}
        \centering
        \caption{Model Performance in Multimodal Generation and Retrieval Tasks. The best result is $\textbf{bold}$. The second result is $\underline{underlined}$.}
        \label{tab:Q5-2}
        \resizebox{\linewidth}{1.7cm}{
            \setlength{\tabcolsep}{3pt} 
            \begin{tabular}{c|cc|cc|cc}
                \toprule
                \multirow{3}{*}{\centering\textbf{Methods}}
                & \multicolumn{2}{c|}{\textbf{Modality Retrieval}}
                & \multicolumn{2}{c|}{\textbf{G2Text}}
                & \multicolumn{2}{c}{\textbf{G2Image}} \\
                \cmidrule{2-7}
                & \multicolumn{2}{c|}{Toys}
                & \multicolumn{2}{c|}{Flickr30k}
                & \multicolumn{2}{c}{SemArt} \\
                & T2I MRR & I2T MRR
                & BLEU-4 & CIDEr
                & CLIP-S & DINO-S \\
                \midrule
                MLP
                & $\textbf{99.46}$ & $\textbf{99.45}$
                & $5.87$ & $\underline{39.37}$
                & $67.54$ & $49.94$ \\
                GCN
                & $96.01$ & $95.96$
                & $5.69$ & $38.44$
                & $67.15$ & $49.65$ \\
                DGF
                & $94.20$ & $94.10$
                & $\textbf{6.83}$ & $\textbf{44.28}$
                & $\underline{68.43}$ & $\underline{52.30}$ \\
                LGMRec
                & $\underline{98.41}$ & $\underline{98.43}$
                & $\underline{5.95}$ & $39.00$
                & $\textbf{68.47}$ & $\textbf{52.73}$ \\
                \bottomrule
            \end{tabular}
        }
    \end{minipage}
    \hfill
    \begin{minipage}{0.45\linewidth}
        \centering
        \caption{Impact of downstream LLM fine-tuning strategies on model generation capabilities.}
        \label{tab:Q6}
        \resizebox{\linewidth}{1.7cm}{
            \begin{tabular}{c|c ccc}
                \toprule
                \textbf{Model} & \textbf{Strategies} & \textbf{BLEU-4} & \textbf{ROUGE-L} & \textbf{CIDEr} \\
                \midrule
                \cmidrule(lr){3-5}
                \multirow{3}{*}{OPT}
                & None   & 6.01 & 29.36 & 41.51 \\
                & Adapter & 5.97 & 27.42 & 34.95 \\
                & LoRA   & 6.63 & 30.00 & 46.23 \\
                \midrule
                \multirow{3}{*}{LLaMA}
                & None   & 5.86 & 28.24 & 44.58 \\
                & Adapter & 6.26 & 28.87 & 42.85 \\
                & LoRA   & 6.20 & 28.85 & 47.76 \\
                \bottomrule
            \end{tabular}
        }
    \end{minipage}
\end{table*}

\begin{figure*}[t]
    \centering
    \includegraphics[width=1.0\linewidth]{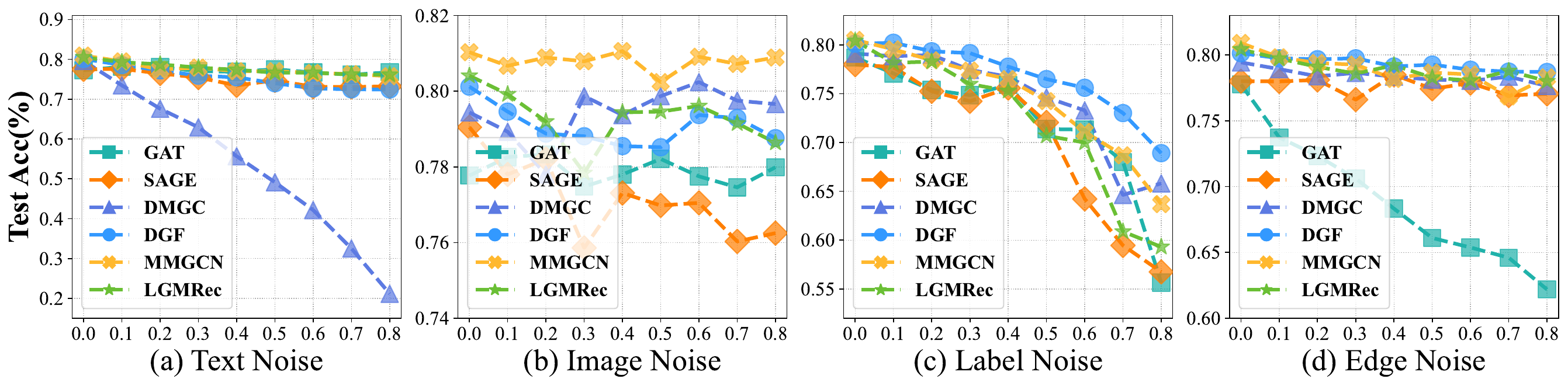} 
    \vspace{-0.3cm} 
    \caption{Robustness analysis on different models.}
    \label{fig:Q7+Q8}
\end{figure*}

\begin{figure*}[t]
    \centering
    \begin{minipage}{0.48\textwidth}
        \centering
        \includegraphics[width=\linewidth]{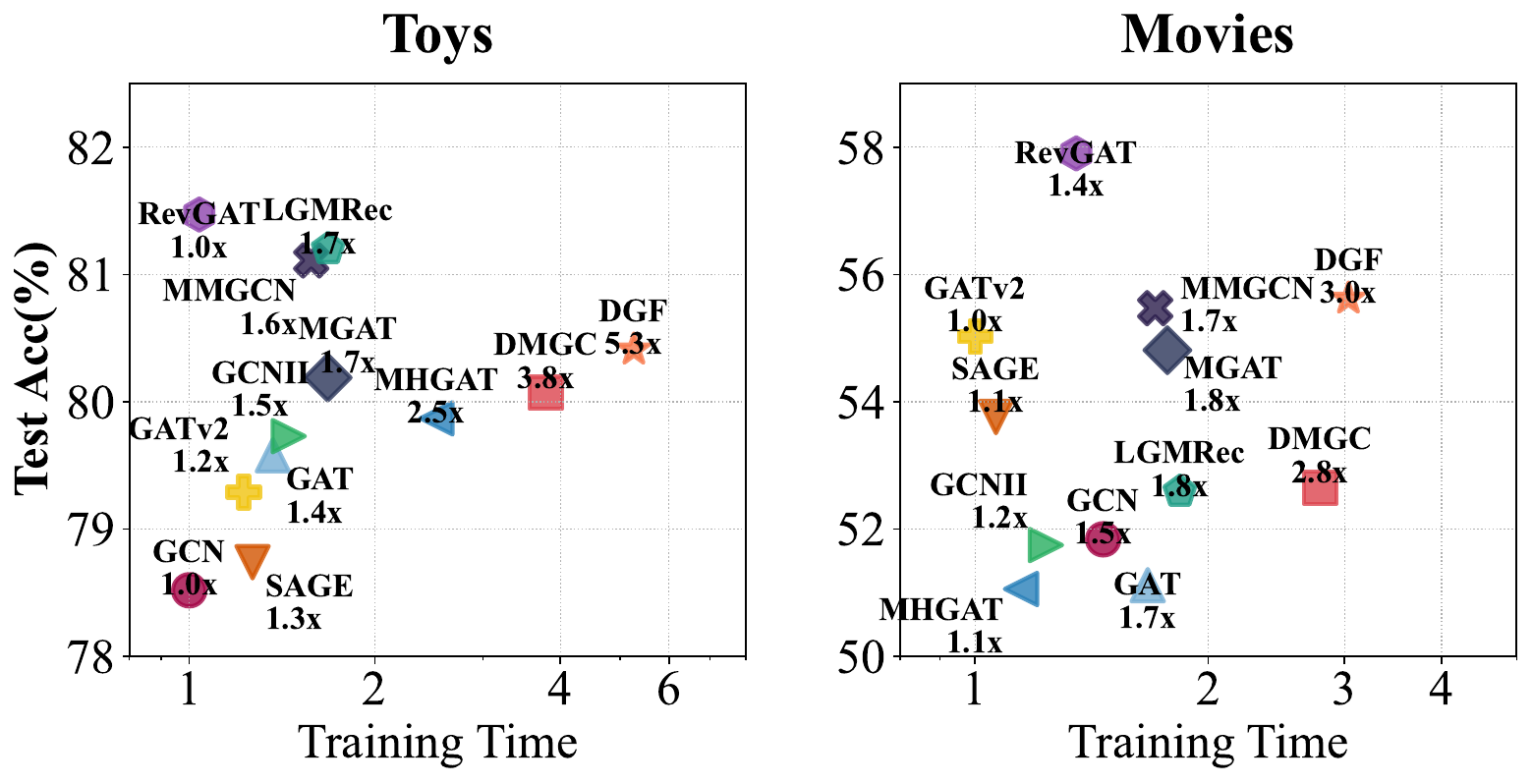}
        \vspace{-0.2cm}
        \caption{Training Time Performance on Various Datasets.}
        \label{fig:Q10_1}
    \end{minipage}
    \hfill
    \begin{minipage}{0.48\textwidth}
        \centering
        \includegraphics[width=\linewidth]{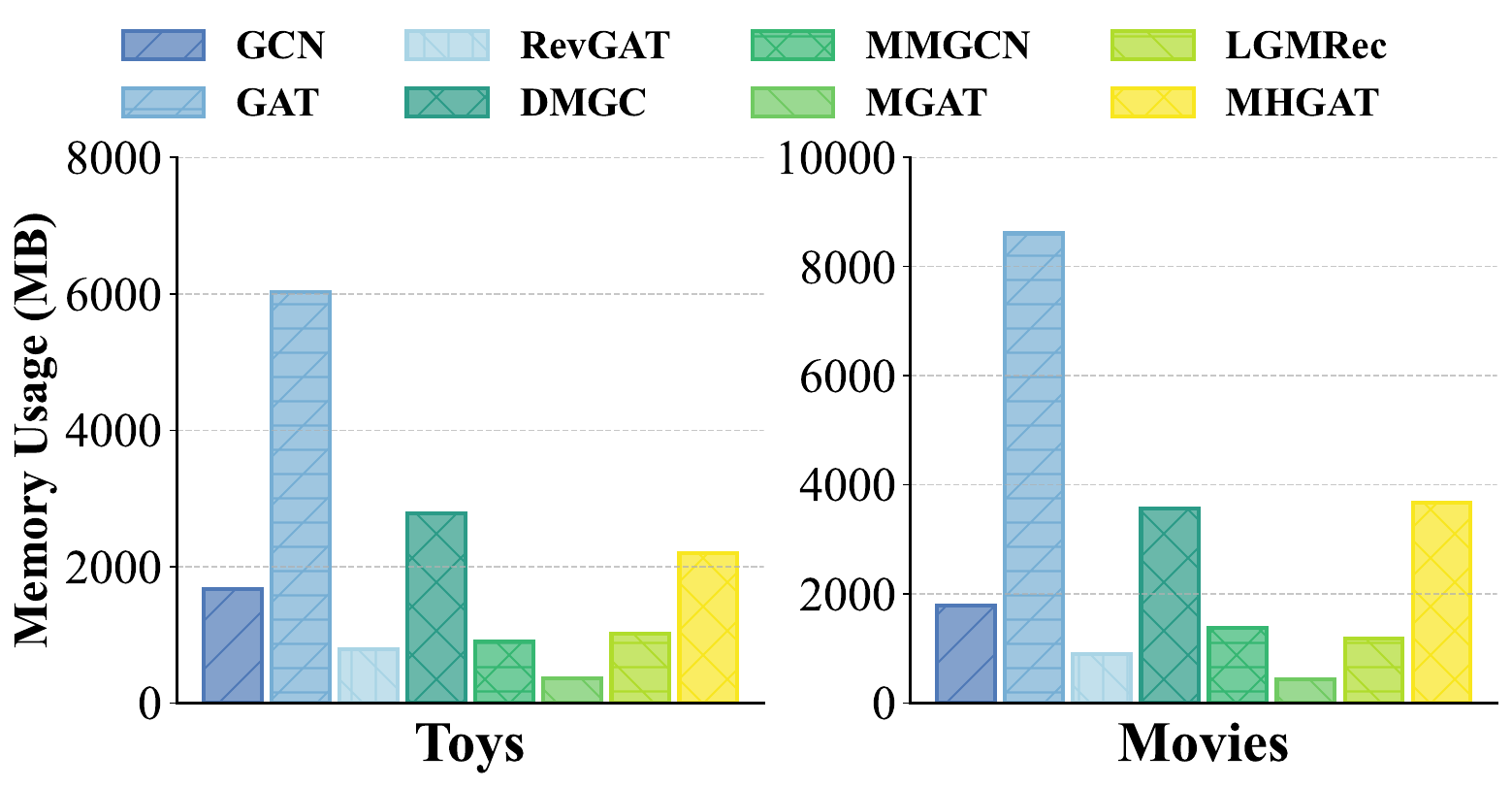}
        \vspace{-0.4cm}
        \caption{Memory Usage Performance on Various Datasets.}
        \label{fig:Q10_2}
    \end{minipage}
    \vspace{-0.15cm}
\end{figure*}

\begin{table}[t]
\centering
\caption{Theoretical complexity analysis of representative models in OpenMAG.}
\label{tab:complexity}
\resizebox{0.48\textwidth}{!}{%
\begin{tabular}{l ccc}
\toprule
\textbf{Models} & \textbf{Training} & \textbf{Inference} & \textbf{Memory} \\
\midrule
GraphMAE2 & $O(L|\mathcal{E}|d + L|\mathcal{V}|d^2)$ & $O(L|\mathcal{E}|d + L|\mathcal{V}|d^2)$ & $O(L|\mathcal{V}|d + |\mathcal{E}|)$ \\
DMGC & $O(|\mathcal{V}|^2 d + K|\mathcal{V}|d)$ & $O(|\mathcal{V}|^2 d)$ & $O(|\mathcal{V}|^2 + K d)$ \\
MIG-GT & $O(L|\mathcal{E}|d + M|\mathcal{V}|d^2)$ & $O(L|\mathcal{E}|d)$ & $O(M|\mathcal{V}|d + |\mathcal{E}|)$ \\
DGF & $O(|\mathcal{E}|d + |\mathcal{V}|d^2)$ & $O(|\mathcal{E}|d)$ & $O(|\mathcal{V}|d + |\mathcal{E}|)$ \\
\midrule
MMGCN & $O(M L |\mathcal{E}| d)$ & $O(M L |\mathcal{E}| d)$ & $O(M L |\mathcal{V}| d)$ \\
MGAT & $O(M L (|\mathcal{V}| d^2 + |\mathcal{E}| d))$ & $O(M L (|\mathcal{V}| d^2 + |\mathcal{E}| d))$ & $O(M L |\mathcal{V}| d + |\mathcal{E}|)$ \\
GSMN & $O(|\mathcal{E}|d + M|\mathcal{V}|^2)$ & $O(|\mathcal{E}|d)$ & $O(M|\mathcal{V}|d + |\mathcal{V}|^2)$ \\
MHGAT & $O(L|\mathcal{E}|d M^2)$ & $O(L|\mathcal{E}|d M^2)$ & $O(M|\mathcal{V}|d + |\mathcal{E}|)$ \\
MGNet & $O(L|\mathcal{E}|d + M|\mathcal{V}|d^2)$ & $O(L|\mathcal{E}|d + M|\mathcal{V}|d^2)$ & $O(M|\mathcal{V}|d + |\mathcal{E}|)$ \\
LGMRec & $O(L|\mathcal{E}|d + M|\mathcal{V}|d)$ & $O(L|\mathcal{E}|d + M|\mathcal{V}|d)$ & $O((M+L)|\mathcal{V}|d + |\mathcal{E}|)$ \\
\midrule
UniGraph2 & $O(L|\mathcal{E}|d + |\mathcal{V}|d^2)$ & $O(L|\mathcal{E}|d)$ & $O(L|\mathcal{V}|d)$ \\
GraphGPT-O & $O(S^2 |\mathcal{V}| + S \Theta)$ & $O(S^2 |\mathcal{V}| + S \Theta)$ & $O(\Theta + S |\mathcal{V}|)$ \\
MLaGA & $O(|\mathcal{E}|d + |\mathcal{V}|d \cdot \Theta_{enc})$ & $O(|\mathcal{E}|d)$ & $O(|\mathcal{V}|d + \Theta_{enc})$ \\
Graph4MM & $O(|\mathcal{E}|d + S^2 |\mathcal{V}|)$ & $O(|\mathcal{E}|d + S |\mathcal{V}|)$ & $O(|\mathcal{V}|d + \Theta)$ \\
NTSFormer & $O(|\mathcal{E}|d + |\mathcal{V}|S^2)$ & $O(|\mathcal{E}|d + |\mathcal{V}|S^2)$ & $O(\Theta + |\mathcal{V}|S)$ \\
InstructG2I & $O(|\mathcal{E}|d + S \Theta)$ & $O(|\mathcal{E}|d + S \Theta)$ & $O(\Theta + |\mathcal{V}|d)$ \\
\bottomrule
\end{tabular}%
}
\vspace{-0.15cm}
\end{table}

\subsection{Assessment of Multimodal Feature Quality}

To address \textbf{Q3}, we examine the necessity of fine-tuning encoders on the node classification task. We compare four settings ranging from fully frozen parameters to fully fine-tuned encoders, utilizing an MLP as the baseline model. As illustrated in Figure~\ref{fig:q3_results}, strategies involving fine-tuning consistently outperform purely frozen baselines. This phenomenon indicates that while pre-trained models encapsulate rich general knowledge, they inevitably suffer from domain shift when applied to specific graph contexts. However, it is worth noting that this performance gain comes at the cost of increased training time and memory usage compared to the highly efficient frozen setting. Therefore, while frozen features offer a low-resource alternative, task-specific adaptation is preferable for scenarios prioritizing maximum performance.

To address \textbf{Q4}, we investigate how different feature encoding strategies impact the semantic alignment of multimodal embeddings on the modality retrieval task using the Flickr30k dataset, with an MLP serving as the baseline model. We contrast Independent Encoders which stitch together separate unimodal models against Unified Encoders jointly pre-trained on image-text pairs (e.g., CLIP, Qwen2.5). The radar chart in Figure~\ref{fig:q4_results} reveals that despite the individual strength of large-scale unimodal models, Independent strategies often suffer from modality misalignment, leading to suboptimal retrieval performance. In contrast, Unified Encoders demonstrate superior cross-modal alignment, suggesting that the synergy established during joint pre-training is critical for retrieval-based tasks.

Based on the observations above, we can conclude that: \textbf{C5}: \textit{Fine-tuning strategies achieve superior downstream performance compared to frozen baselines by mitigating domain shifts, albeit with higher computational overhead.}~\cite{Conclusion5} \textbf{C6}: \textit{Unified encoders jointly pre-trained on multimodal data provide fundamentally better semantic alignment and feature quality than combinations of independent unimodal encoders.}~\cite{clip}

\subsection{Effectiveness Analysis}
To address Q5, we benchmark diverse models classified into Traditional GNNs, Graph-enhanced methods, Multimodal-enhanced frameworks, and MLLM-enhanced models. We also incorporate an MLP as a simple baseline. As shown in Table~\ref{tab:Q5_1}, for Graph-Based tasks, LGMRec and DGF demonstrated superior performance over traditional baselines. This dominance indicates that relying solely on topology is insufficient; instead, advanced structural denoising and the simultaneous modeling of cross-modal correlations are crucial for capturing complex dependencies in these scenarios. UniGraph2 exhibited suboptimal performance under these specific settings. This outcome likely stems from the inherent limitations of its generative masked-prediction paradigm, which focuses on reconstruction, and thus may lack the direct discriminative power of specialized contrastive objectives that are optimized for specific structural tasks. For Modality-Based tasks, we evaluate model performance across semantic alignment and generative scenarios. Specifically, we select Modality Retrieval to represent the semantic alignment tasks. As shown in Table~\ref{tab:Q5-2}, MLP outperforms other models in retrieval, implying that structural aggregation might introduce noise. For generative tasks, DGF outperforms others in G2Text, indicating that structural denoising benefits captioning. Conversely, LGMRec slightly outperforms others, suggesting that modeling cross-modal correlations is essential for high-fidelity visual synthesis.

To address \textbf{Q6}, we investigate the impact of Parameter-Efficient Fine-Tuning (PEFT) strategies on the generative capabilities of LLMs. We utilize the G2Text task on Flickr30k to evaluate the performance of OPT and LLaMA. Table~\ref{tab:Q6} reveals that while Adapters improve lexical matching, they compromise semantic grounding compared to frozen baselines, likely due to interference with pre-trained representations. In contrast, LoRA emerge as the superior strategy, effectively bridging the modality gap by enabling adaptive focus on graph-aligned tokens. This demonstrates that LoRA provides the optimal lightweight adaptation to align LLM knowledge with MAG patterns without the semantic degradation observed in standard Adapters. Detailed fine-tuning configurations are provided in Appendix~\ref{app:task_hyperparameters}.

Through our experiments above, we derive four key conclusions. \textbf{C7}: \textit{Graph-enhanced models dominate most scenarios, demonstrating structural denoising outperforms pure topology.}~\cite{DGF} \textbf{C8}: \textit{Multimodal-enhanced models excel in node classification and generative tasks, while simple MLP outperform complex architectures in modality retrieval tasks.}~\cite{LGMRec} \textbf{C9}: \textit{MLLM-enhanced paradigms exhibit suboptimal performance in discriminative settings due to the intrinsic objective gap between generative pre-training and structural reasoning.}~\cite{UniGraph2} \textbf{C10}: \textit{LoRA demonstrates superior effectiveness among PEFT strategies and achieve the highest generation quality scores across diverse architectures.}~\cite{lora}

\subsection{Robustness Analysis}

To address \textbf{Q7}, we conducted a systematic evaluation on the Toys dataset to assess model resilience against semantic and label noise within the node classification task. We simulate modality noise by randomly shuffling text or image features to sever semantic alignment, and introduce label noise by randomly flipping ground-truth labels. As observed in Figure \ref{fig:Q7+Q8} (a-c), distinct vulnerability patterns emerge. Under text noise, while most architectures maintain stability, DMGC suffers a catastrophic collapse, indicating its filtering mechanism is overly dependent on textual semantic quality. For label noise, while all models exhibit performance degradation as noise intensity increases, Graph-enhanced models demonstrate significantly stronger robustness compared to standard GNN baselines, suggesting that leveraging auxiliary self-supervised signals or multimodal correlations can effectively buffer against label corruption. 

To address \textbf{Q8}, we investigate model resilience against topological perturbations by randomly adding spurious edges to the original graph structure, simulating the presence of false interactions often encountered in real-world data collection. As shown in Figure \ref{fig:Q7+Q8}d, GAT experiences a sharp decline, confirming that attention mechanisms can easily overfit spurious connections. In contrast, SAGE and Multimodal models remain stable, indicating that aggregating neighbor information via fixed weights or leveraging rich node content reduces sensitivity to structual perturbations.

Based on the observations above, we derive two key conclusions. \textbf{C11}: \textit{Adaptive multimodal integration and self-supervised signals serve as critical stabilizers, allowing models to mitigate the impact of single-modality corruption or label noise where purely supervised approaches fail.} \textbf{C12}: \textit{Robustness against structural noise requires decoupling representation learning from strict reliance on local edge reliability; models that leverage intrinsic node attributes or structure-agnostic components can effectively counteract the interference of spurious topology.}

\subsection{Complexity and Practical Efficiency Analysis}

To address \textbf{Q9}, we analyze the theoretical complexity of MAG models regarding training, inference, and memory footprint. As detailed in Table~\ref{tab:complexity}, graph-enhanced models generally maintain linear scalability relative to edges, although certain methods involving structure learning may introduce quadratic terms $O(|\mathcal{V}|^2)$ due to dense similarity computations. Multimodal-enhanced models inherit the efficiency of standard GNNs but are predominantly governed by the number of modalities $M$, where multimodal attention mechanisms can further increase the computational constant. In contrast, MLLM-enhanced models exhibit significantly higher complexity, where reliance on Transformers introduces quadratic costs $O(S^2)$ relative to sequence length, and the massive parameter scale $\Theta$ imposes substantial memory requirements. Detailed descriptions of experimental details and complexity formulas are provided in Appendix~\ref{app:hyperparameters} and~\ref{app:efficiency}.

To address \textbf{Q10}, we evaluate empirical computational burdens regarding training latency and memory usage. As illustrated in Figure~\ref{fig:Q10_1}, Traditional GNNs consistently exhibit the lowest latency, establishing a highly efficient baseline. In contrast, Graph-enhanced methods incur substantial time overheads attributed to the computational intensity of iterative structure learning modules. Multimodal-enhanced models maintain moderate training costs while achieving competitive accuracy. Regarding memory usage in Figure~\ref{fig:Q10_2}, explicit multi-view graph construction in models such as MMGCN creates severe bottlenecks, whereas LGMRec maintains a low memory usage compared to standard GNNs, highlighting its robustness for large-scale applications. This indicates that architectural efficiency is as critical as accuracy for practical deployment in resource-limited scenarios.

Based on the analysis, we conclude \textbf{C13}: \textit{Graph-enhanced and MLLM-based models are often constrained by quadratic complexity in structure learning or sequence processing, whereas Multimodal-enhanced frameworks generally maintain linear scalability.} \textbf{C14}: \textit{Empirically, Multimodal-enhanced models demonstrate the most favorable trade-off between accuracy and efficiency, while Graph-enhanced models require optimization to mitigate latency and memory overheads in practical applications.}

\section{CONCLUSION AND FUTURE DIRECTIONS}

In this paper, we propose OpenMAG, a comprehensive benchmark for Multimodal-Attributed Graph that integrates 19 datasets across 6 domains, 16 modality encoders, 24 state-of-the-art models, and 8 downstream tasks. Through systematic experiments regarding Necessity, Data Quality, Effectiveness, Robustness, and Efficiency, we derived 14 fundamental insights into MAG learning. Based on these findings, we summarize the critical challenges faced by MAG and potential future directions for the community.

\textbf{Unification of Multimodal and Graph Structures.} \textit{(C1-C4)} Our results confirm that neither unimodal semantics nor pure topology suffices alone; optimal performance demands their deep integration. The core challenge lies in the rigid nature of current fusion mechanisms, which often treat modalities and structure as separate entities. Future research should focus on unified architectures that inherently blend multimodal semantics with structural guidance to fully exploit their complementary necessity.

\textbf{Flexible and Aligned Modality Encoding.} \textit{(C5-C6)} We identify that flexible fine-tuning and precise cross-modal alignment are indispensable for generating high-quality node embeddings. However, achieving such semantic alignment often incurs prohibitive computational costs. Future research should investigate lightweight alignment techniques and more flexible multimodal encoding processes that prioritize precise semantic alignment.

\textbf{Unified Generalist MAG Models.} \textit{(C7-C10)} Current MAG models are highly specialized, excelling in isolated tasks but failing to generalize. This fragmentation limits versatility, as distinct architectures are typically required for different learning objectives. Future work should prioritize Unified MAG Models, developing versatile frameworks capable of adaptively handling diverse structural semantics and downstream scenarios effectively.

\textbf{Robustness in Realistic Scenarios.} \textit{(C11-C12)} Real-world deployments inevitably face semantic corruption and structural unreliability. Our analysis reveals that models relying strictly on local message passing are fragile in such realistic scenarios. Future models must explore structure-agnostic pathways and self-supervised denoising objectives to decouple representation learning from input imperfections, ensuring stability under dual noisy conditions.

\textbf{Scalable and Efficient Architectures.} \textit{(C13-C14)} Existing MAG models are often constrained by quadratic complexity, creating a deployment bottleneck. Balancing between reasoning capabilities and feasibility is critical. Future research must prioritize scalable models that maintain efficiency on large-scale graphs, and explore model compression to transfer heavy capabilities to lightweight counterparts.

\section*{Impact Statement}
This paper presents work whose goal is to advance the field of Machine
Learning.
There are many potential societal consequences of our work, none of which we feel must be specifically highlighted here.

\bibliography{reference}

@inproceedings{mmgraph,
  author       = {Jing Zhu and
                  Yuhang Zhou and
                  Shengyi Qian and
                  Zhongmou He and
                  Tong Zhao and
                  Neil Shah and
                  Danai Koutra},
  title        = {Mosaic of Modalities: {A} Comprehensive Benchmark for Multimodal Graph
                  Learning},
  booktitle    = {{CVPR}},
  pages        = {14215--14224},
  publisher    = {Computer Vision Foundation / {IEEE}},
  year         = {2025}
}

@article{MAGB,
  author       = {Hao Yan and
                  Chaozhuo Li and
                  Zhigang Yu and
                  Jun Yin and
                  Ruochen Liu and
                  Peiyan Zhang and
                  Weihao Han and
                  Mingzheng Li and
                  Zhengxin Zeng and
                  Hao Sun and
                  Weiwei Deng and
                  Feng Sun and
                  Qi Zhang and
                  Senzhang Wang},
  title        = {When Graph meets Multimodal: Benchmarking on Multimodal Attributed
                  Graphs Learning},
  journal      = {CoRR},
  volume       = {abs/2410.09132},
  year         = {2024}
}

@inproceedings{DBLP:conf/www/Fan0LHZTY19,
  author       = {Wenqi Fan and
                  Yao Ma and
                  Qing Li and
                  Yuan He and
                  Yihong Eric Zhao and
                  Jiliang Tang and
                  Dawei Yin},
  title        = {Graph Neural Networks for Social Recommendation},
  booktitle    = {{WWW}},
  pages        = {417--426},
  publisher    = {{ACM}},
  year         = {2019}
}

@inproceedings{MMGCN,
  author       = {Yinwei Wei and
                  Xiang Wang and
                  Liqiang Nie and
                  Xiangnan He and
                  Richang Hong and
                  Tat{-}Seng Chua},
  title        = {{MMGCN:} Multi-modal Graph Convolution Network for Personalized Recommendation
                  of Micro-video},
  booktitle    = {{ACM} Multimedia},
  pages        = {1437--1445},
  publisher    = {{ACM}},
  year         = {2019}
}

@inproceedings{DBLP:conf/aaai/CaiWLZ024,
  author       = {Jie Cai and
                  Xin Wang and
                  Haoyang Li and
                  Ziwei Zhang and
                  Wenwu Zhu},
  title        = {Multimodal Graph Neural Architecture Search under Distribution Shifts},
  booktitle    = {{AAAI}},
  pages        = {8227--8235},
  publisher    = {{AAAI} Press},
  year         = {2024}
}

@article{DBLP:journals/corr/abs-2402-05322,
  author       = {Ciyuan Peng and
                  Jiayuan He and
                  Feng Xia},
  title        = {Learning on Multimodal Graphs: {A} Survey},
  journal      = {CoRR},
  volume       = {abs/2402.05322},
  year         = {2024}
}

@inproceedings{GCN,
  author       = {Thomas N. Kipf and
                  Max Welling},
  title        = {Semi-Supervised Classification with Graph Convolutional Networks},
  booktitle    = {{ICLR} (Poster)},
  publisher    = {OpenReview.net},
  year         = {2017}
}

@inproceedings{
GAT,
title={Graph Attention Networks},
author={Petar Veličković and Guillem Cucurull and Arantxa Casanova and Adriana Romero and Pietro Liò and Yoshua Bengio},
booktitle={International Conference on Learning Representations},
year={2018}
}

@article{MGAT,
  author       = {Zhulin Tao and
                  Yinwei Wei and
                  Xiang Wang and
                  Xiangnan He and
                  Xianglin Huang and
                  Tat{-}Seng Chua},
  title        = {{MGAT:} Multimodal Graph Attention Network for Recommendation},
  journal      = {Inf. Process. Manag.},
  volume       = {57},
  number       = {5},
  pages        = {102277},
  year         = {2020}
}

@inproceedings{UniGraph2,
  author       = {Yufei He and
                  Yuan Sui and
                  Xiaoxin He and
                  Yue Liu and
                  Yifei Sun and
                  Bryan Hooi},
  title        = {UniGraph2: Learning a Unified Embedding Space to Bind Multimodal Graphs},
  booktitle    = {{WWW}},
  pages        = {1759--1770},
  publisher    = {{ACM}},
  year         = {2025}
}

@inproceedings{GraphGPT-O,
  author       = {Yi Fang and
                  Bowen Jin and
                  Jiacheng Shen and
                  Sirui Ding and
                  Qiaoyu Tan and
                  Jiawei Han},
  title        = {{GRAPHGPT-O:} Synergistic Multimodal Comprehension and Generation
                  on Graphs},
  booktitle    = {{CVPR}},
  pages        = {19467--19476},
  publisher    = {Computer Vision Foundation / {IEEE}},
  year         = {2025}
}

@article{ninerec,
  author       = {Jiaqi Zhang and
                  Yu Cheng and
                  Yongxin Ni and
                  Yunzhu Pan and
                  Zheng Yuan and
                  Junchen Fu and
                  Youhua Li and
                  Jie Wang and
                  Fajie Yuan},
  title        = {NineRec: {A} Benchmark Dataset Suite for Evaluating Transferable Recommendation},
  journal      = {{IEEE} Trans. Pattern Anal. Mach. Intell.},
  volume       = {47},
  number       = {7},
  pages        = {5256--5267},
  year         = {2025}
}

@inproceedings{sbert,
  author       = {Nils Reimers and
                  Iryna Gurevych},
  title        = {Sentence-BERT: Sentence Embeddings using Siamese BERT-Networks},
  booktitle    = {{EMNLP/IJCNLP} {(1)}},
  pages        = {3980--3990},
  publisher    = {Association for Computational Linguistics},
  year         = {2019}
}

@article{llama3.2,
  author       = {Llama Team},
  title        = {The Llama 3 Herd of Models},
  journal      = {CoRR},
  volume       = {abs/2407.21783},
  year         = {2024}
}

@article{vit,
  author       = {Bichen Wu and
                  Chenfeng Xu and
                  Xiaoliang Dai and
                  Alvin Wan and
                  Peizhao Zhang and
                  Masayoshi Tomizuka and
                  Kurt Keutzer and
                  Peter Vajda},
  title        = {Visual Transformers: Token-based Image Representation and Processing
                  for Computer Vision},
  journal      = {CoRR},
  volume       = {abs/2006.03677},
  year         = {2020}
}

@inproceedings{imagenet,
  author       = {Jia Deng and
                  Wei Dong and
                  Richard Socher and
                  Li{-}Jia Li and
                  Kai Li and
                  Li Fei{-}Fei},
  title        = {ImageNet: {A} large-scale hierarchical image database},
  booktitle    = {{CVPR}},
  pages        = {248--255},
  publisher    = {{IEEE} Computer Society},
  year         = {2009}
}

@inproceedings{swin,
  author       = {Ze Liu and
                  Yutong Lin and
                  Yue Cao and
                  Han Hu and
                  Yixuan Wei and
                  Zheng Zhang and
                  Stephen Lin and
                  Baining Guo},
  title        = {Swin Transformer: Hierarchical Vision Transformer using Shifted Windows},
  booktitle    = {{ICCV}},
  pages        = {9992--10002},
  publisher    = {{IEEE}},
  year         = {2021}
}

@inproceedings{clip,
  author       = {Alec Radford and
                  Jong Wook Kim and
                  Chris Hallacy and
                  Aditya Ramesh and
                  Gabriel Goh and
                  Sandhini Agarwal and
                  Girish Sastry and
                  Amanda Askell and
                  Pamela Mishkin and
                  Jack Clark and
                  Gretchen Krueger and
                  Ilya Sutskever},
  title        = {Learning Transferable Visual Models From Natural Language Supervision},
  booktitle    = {{ICML}},
  series       = {Proceedings of Machine Learning Research},
  volume       = {139},
  pages        = {8748--8763},
  publisher    = {{PMLR}},
  year         = {2021}
}

@misc{qwenvl,
      title={Qwen-VL: A Versatile Vision-Language Model for Understanding, Localization, Text Reading, and Beyond}, 
      author={Jinze Bai and Shuai Bai and Shusheng Yang and Shijie Wang and Sinan Tan and Peng Wang and Junyang Lin and Chang Zhou and Jingren Zhou},
      year={2023},
      eprint={2308.12966},
      archivePrefix={arXiv},
      primaryClass={cs.CV}
}

@article{GraphMLLM,
  author       = {Jiajin Liu and
                  Dongzhe Fan and
                  Jiacheng Shen and
                  Chuanhao Ji and
                  Daochen Zha and
                  Qiaoyu Tan},
  title        = {Graph-MLLM: Harnessing Multimodal Large Language Models for Multimodal
                  Graph Learning},
  journal      = {CoRR},
  volume       = {abs/2506.10282},
  year         = {2025}
}

@article{DMGC,
  author       = {Zhaochen Guo and
                  Zhixiang Shen and
                  Xuanting Xie and
                  Liangjian Wen and
                  Zhao Kang},
  title        = {Disentangling Homophily and Heterophily in Multimodal Graph Clustering},
  journal      = {CoRR},
  volume       = {abs/2507.15253},
  year         = {2025}
}

@misc{DGF,
      title={Cross-Contrastive Clustering for Multimodal Attributed Graphs with Dual Graph Filtering}, 
      author={Haoran Zheng and Renchi Yang and Hongtao Wang and Jianliang Xu},
      year={2025},
      eprint={2511.20030},
      archivePrefix={arXiv},
      primaryClass={cs.LG}
}

@article{MLaGA,
  author       = {Dongzhe Fan and
                  Yi Fang and
                  Jiajin Liu and
                  Djellel Difallah and
                  Qiaoyu Tan},
  title        = {MLaGA: Multimodal Large Language and Graph Assistant},
  journal      = {CoRR},
  volume       = {abs/2506.02568},
  year         = {2025}
}

@inproceedings{GraphSAGE,
  author       = {William L. Hamilton and
                  Zhitao Ying and
                  Jure Leskovec},
  title        = {Inductive Representation Learning on Large Graphs},
  booktitle    = {{NIPS}},
  pages        = {1024--1034},
  year         = {2017}
}

@inproceedings{LGMRec,
  author       = {Zhiqiang Guo and
                  Jianjun Li and
                  Guohui Li and
                  Chaoyang Wang and
                  Si Shi and
                  Bin Ruan},
  title        = {LGMRec: Local and Global Graph Learning for Multimodal Recommendation},
  booktitle    = {{AAAI}},
  pages        = {8454--8462},
  publisher    = {{AAAI} Press},
  year         = {2024}
}

@inproceedings{Amazon2018,
  author       = {Jianmo Ni and
                  Jiacheng Li and
                  Julian J. McAuley},
  title        = {Justifying Recommendations using Distantly-Labeled Reviews and Fine-Grained
                  Aspects},
  booktitle    = {{EMNLP/IJCNLP} {(1)}},
  pages        = {188--197},
  publisher    = {Association for Computational Linguistics},
  year         = {2019}
}

@inproceedings{RedditS,
  author       = {Karan Desai and
                  Gaurav Kaul and
                  Zubin Aysola and
                  Justin Johnson},
  title        = {RedCaps: Web-curated image-text data created by the people, for the
                  people},
  booktitle    = {NeurIPS Datasets and Benchmarks},
  year         = {2021}
}

@misc{Amazonreview1,
      title={Bridging Language and Items for Retrieval and Recommendation}, 
      author={Yupeng Hou and Jiacheng Li and Zhankui He and An Yan and Xiusi Chen and Julian McAuley},
      year={2024},
      eprint={2403.03952},
      archivePrefix={arXiv},
      primaryClass={cs.IR},
}

@inproceedings{Amazonreview2,
  author       = {Jianmo Ni and
                  Jiacheng Li and
                  Julian J. McAuley},
  title        = {Justifying Recommendations using Distantly-Labeled Reviews and Fine-Grained
                  Aspects},
  booktitle    = {{EMNLP/IJCNLP} {(1)}},
  pages        = {188--197},
  publisher    = {Association for Computational Linguistics},
  year         = {2019}
}

@inproceedings{books-nc1,
  author       = {Mengting Wan and
                  Julian J. McAuley},
  title        = {Item recommendation on monotonic behavior chains},
  booktitle    = {RecSys},
  pages        = {86--94},
  publisher    = {{ACM}},
  year         = {2018}
}

@inproceedings{books-nc2,
  author       = {Mengting Wan and
                  Rishabh Misra and
                  Ndapa Nakashole and
                  Julian J. McAuley},
  title        = {Fine-Grained Spoiler Detection from Large-Scale Review Corpora},
  booktitle    = {{ACL} {(1)}},
  pages        = {2605--2610},
  publisher    = {Association for Computational Linguistics},
  year         = {2019}
}

@inproceedings{flickr30k,
  author       = {Bryan A. Plummer and
                  Liwei Wang and
                  Chris M. Cervantes and
                  Juan C. Caicedo and
                  Julia Hockenmaier and
                  Svetlana Lazebnik},
  title        = {Flickr30k Entities: Collecting Region-to-Phrase Correspondences for
                  Richer Image-to-Sentence Models},
  booktitle    = {{ICCV}},
  pages        = {2641--2649},
  publisher    = {{IEEE} Computer Society},
  year         = {2015}
}

@inproceedings{semart,
  author       = {Noa Garcia and
                  George Vogiatzis},
  title        = {How to Read Paintings: Semantic Art Understanding with Multi-modal
                  Retrieval},
  booktitle    = {{ECCV} Workshops {(2)}},
  series       = {Lecture Notes in Computer Science},
  volume       = {11130},
  pages        = {676--691},
  publisher    = {Springer},
  year         = {2018}
}

@inproceedings{convnextv2,
  author       = {Sanghyun Woo and
                  Shoubhik Debnath and
                  Ronghang Hu and
                  Xinlei Chen and
                  Zhuang Liu and
                  In So Kweon and
                  Saining Xie},
  title        = {ConvNeXt {V2:} Co-designing and Scaling ConvNets with Masked Autoencoders},
  booktitle    = {{CVPR}},
  pages        = {16133--16142},
  publisher    = {{IEEE}},
  year         = {2023}
}

@article{dinov2,
  author       = {Maxime Oquab and
                  Timoth{\'{e}}e Darcet and
                  Th{\'{e}}o Moutakanni and
                  Huy V. Vo and
                  Marc Szafraniec and
                  Vasil Khalidov and
                  Pierre Fernandez and
                  Daniel Haziza and
                  Francisco Massa and
                  Alaaeldin El{-}Nouby and
                  Mido Assran and
                  Nicolas Ballas and
                  Wojciech Galuba and
                  Russell Howes and
                  Po{-}Yao Huang and
                  Shang{-}Wen Li and
                  Ishan Misra and
                  Michael Rabbat and
                  Vasu Sharma and
                  Gabriel Synnaeve and
                  Hu Xu and
                  Herv{\'{e}} J{\'{e}}gou and
                  Julien Mairal and
                  Patrick Labatut and
                  Armand Joulin and
                  Piotr Bojanowski},
  title        = {DINOv2: Learning Robust Visual Features without Supervision},
  journal      = {Trans. Mach. Learn. Res.},
  volume       = {2024},
  year         = {2024}
}

@inproceedings{vig,
  author       = {Kai Han and
                  Yunhe Wang and
                  Jianyuan Guo and
                  Yehui Tang and
                  Enhua Wu},
  title        = {Vision {GNN:} An Image is Worth Graph of Nodes},
  booktitle    = {NeurIPS},
  year         = {2022}
}

@misc{tnt,
      title={Transformer in Transformer}, 
      author={Kai Han and An Xiao and Enhua Wu and Jianyuan Guo and Chunjing Xu and Yunhe Wang},
      year={2021},
      eprint={2103.00112},
      archivePrefix={arXiv},
      primaryClass={cs.CV}
}

@misc{roberta,
      title={RoBERTa: A Robustly Optimized BERT Pretraining Approach}, 
      author={Yinhan Liu and Myle Ott and Naman Goyal and Jingfei Du and Mandar Joshi and Danqi Chen and Omer Levy and Mike Lewis and Luke Zettlemoyer and Veselin Stoyanov},
      year={2019},
      eprint={1907.11692},
      archivePrefix={arXiv},
      primaryClass={cs.CL},
}

@misc{opt,
      title={OPT: Open Pre-trained Transformer Language Models}, 
      author={Susan Zhang and Stephen Roller and Naman Goyal and Mikel Artetxe and Moya Chen and Shuohui Chen and Christopher Dewan and Mona Diab and Xian Li and Xi Victoria Lin and Todor Mihaylov and Myle Ott and Sam Shleifer and Kurt Shuster and Daniel Simig and Punit Singh Koura and Anjali Sridhar and Tianlu Wang and Luke Zettlemoyer},
      year={2022},
      eprint={2205.01068},
      archivePrefix={arXiv},
      primaryClass={cs.CL}
}

@article{t5,
  author       = {Colin Raffel and
                  Noam Shazeer and
                  Adam Roberts and
                  Katherine Lee and
                  Sharan Narang and
                  Michael Matena and
                  Yanqi Zhou and
                  Wei Li and
                  Peter J. Liu},
  title        = {Exploring the Limits of Transfer Learning with a Unified Text-to-Text
                  Transformer},
  journal      = {J. Mach. Learn. Res.},
  volume       = {21},
  pages        = {140:1--140:67},
  year         = {2020}
}

@inproceedings{imagebind,
  author       = {Rohit Girdhar and
                  Alaaeldin El{-}Nouby and
                  Zhuang Liu and
                  Mannat Singh and
                  Kalyan Vasudev Alwala and
                  Armand Joulin and
                  Ishan Misra},
  title        = {ImageBind One Embedding Space to Bind Them All},
  booktitle    = {{CVPR}},
  pages        = {15180--15190},
  publisher    = {{IEEE}},
  year         = {2023}
}

@inproceedings{revgat,
  author       = {Guohao Li and
                  Matthias M{\"{u}}ller and
                  Bernard Ghanem and
                  Vladlen Koltun},
  title        = {Training Graph Neural Networks with 1000 Layers},
  booktitle    = {{ICML}},
  series       = {Proceedings of Machine Learning Research},
  volume       = {139},
  pages        = {6437--6449},
  publisher    = {{PMLR}},
  year         = {2021}
}

@inproceedings{gin ,
  author       = {Keyulu Xu and
                  Weihua Hu and
                  Jure Leskovec and
                  Stefanie Jegelka},
  title        = {How Powerful are Graph Neural Networks?},
  booktitle    = {{ICLR}},
  publisher    = {OpenReview.net},
  year         = {2019}
}

@misc{chebnet,
      title={ChebNet: Efficient and Stable Constructions of Deep Neural Networks with Rectified Power Units via Chebyshev Approximations}, 
      author={Shanshan Tang and Bo Li and Haijun Yu},
      year={2023},
      eprint={1911.05467},
      archivePrefix={arXiv},
      primaryClass={cs.LG}
}

@inproceedings{gcn2,
  author       = {Ming Chen and
                  Zhewei Wei and
                  Zengfeng Huang and
                  Bolin Ding and
                  Yaliang Li},
  title        = {Simple and Deep Graph Convolutional Networks},
  booktitle    = {{ICML}},
  series       = {Proceedings of Machine Learning Research},
  volume       = {119},
  pages        = {1725--1735},
  publisher    = {{PMLR}},
  year         = {2020}
}

@inproceedings{gat2,
  author       = {Shaked Brody and
                  Uri Alon and
                  Eran Yahav},
  title        = {How Attentive are Graph Attention Networks?},
  booktitle    = {{ICLR}},
  publisher    = {OpenReview.net},
  year         = {2022}
}

@inproceedings{gsmn,
  author       = {Chunxiao Liu and
                  Zhendong Mao and
                  Tianzhu Zhang and
                  Hongtao Xie and
                  Bin Wang and
                  Yongdong Zhang},
  title        = {Graph Structured Network for Image-Text Matching},
  booktitle    = {{CVPR}},
  pages        = {10918--10927},
  publisher    = {Computer Vision Foundation / {IEEE}},
  year         = {2020}
}

@article{mgnet,
  author       = {Zhaoming Kong and
                  Lichao Sun and
                  Hao Peng and
                  Liang Zhan and
                  Yong Chen and
                  Lifang He},
  title        = {Multiplex Graph Networks for Multimodal Brain Network Analysis},
  journal      = {CoRR},
  volume       = {abs/2108.00158},
  year         = {2021}
}

@article{mhgat,
  author       = {Xiangen Jia and
                  Min Jiang and
                  Yihong Dong and
                  Feng Zhu and
                  Haocai Lin and
                  Yu Xin and
                  Huahui Chen},
  title        = {Multimodal heterogeneous graph attention network},
  journal      = {Neural Comput. Appl.},
  volume       = {35},
  number       = {4},
  pages        = {3357--3372},
  year         = {2023}
}

@inproceedings{graphmae2,
  author       = {Zhenyu Hou and
                  Yufei He and
                  Yukuo Cen and
                  Xiao Liu and
                  Yuxiao Dong and
                  Evgeny Kharlamov and
                  Jie Tang},
  title        = {GraphMAE2: {A} Decoding-Enhanced Masked Self-Supervised Graph Learner},
  booktitle    = {{WWW}},
  pages        = {737--746},
  publisher    = {{ACM}},
  year         = {2023}
}

@inproceedings{MIG-GT,
  author       = {Jun Hu and
                  Bryan Hooi and
                  Bingsheng He and
                  Yinwei Wei},
  title        = {Modality-Independent Graph Neural Networks with Global Transformers
                  for Multimodal Recommendation},
  booktitle    = {{AAAI}},
  pages        = {11790--11798},
  publisher    = {{AAAI} Press},
  year         = {2025}
}

@article{NTSFormer,
  author       = {Jun Hu and
                  Yufei He and
                  Yuan Li and
                  Bryan Hooi and
                  Bingsheng He},
  title        = {NTSFormer: {A} Self-Teaching Graph Transformer for Multimodal Cold-Start
                  Node Classification},
  journal      = {CoRR},
  volume       = {abs/2507.04870},
  year         = {2025}
}

@inproceedings{graph4mm,
  author       = {Xuying Ning and
                  Dongqi Fu and
                  Tianxin Wei and
                  Wujiang Xu and
                  Jingrui He},
  title        = {Graph4MM: Weaving Multimodal Learning with Structural Information},
  booktitle    = {{ICML}},
  publisher    = {OpenReview.net},
  year         = {2025}
}

@inproceedings{instructg2i,
  author       = {Bowen Jin and
                  Ziqi Pang and
                  Bingjun Guo and
                  Yu{-}Xiong Wang and
                  Jiaxuan You and
                  Jiawei Han},
  title        = {InstructG2I: Synthesizing Images from Multimodal Attributed Graphs},
  booktitle    = {NeurIPS},
  year         = {2024}
}

@misc{MMGL,
      title={Multimodal Graph Learning for Generative Tasks}, 
      author={Minji Yoon and Jing Yu Koh and Bryan Hooi and Ruslan Salakhutdinov},
      year={2023},
      eprint={2310.07478},
      archivePrefix={arXiv},
      primaryClass={cs.AI}
}

@inproceedings{Conclusion5,
  author       = {Suchin Gururangan and
                  Ana Marasovic and
                  Swabha Swayamdipta and
                  Kyle Lo and
                  Iz Beltagy and
                  Doug Downey and
                  Noah A. Smith},
  title        = {Don't Stop Pretraining: Adapt Language Models to Domains and Tasks},
  booktitle    = {{ACL}},
  pages        = {8342--8360},
  publisher    = {Association for Computational Linguistics},
  year         = {2020}
}

@inproceedings{lora,
  author       = {Edward J. Hu and
                  Yelong Shen and
                  Phillip Wallis and
                  Zeyuan Allen{-}Zhu and
                  Yuanzhi Li and
                  Shean Wang and
                  Lu Wang and
                  Weizhu Chen},
  title        = {LoRA: Low-Rank Adaptation of Large Language Models},
  booktitle    = {{ICLR}},
  publisher    = {OpenReview.net},
  year         = {2022}
}

@inproceedings{adamw,
  author       = {Ilya Loshchilov and
                  Frank Hutter},
  title        = {Decoupled Weight Decay Regularization},
  booktitle    = {{ICLR} (Poster)},
  publisher    = {OpenReview.net},
  year         = {2019}
}

@inproceedings{ia3,
  author       = {Haokun Liu and
                  Derek Tam and
                  Mohammed Muqeeth and
                  Jay Mohta and
                  Tenghao Huang and
                  Mohit Bansal and
                  Colin Raffel},
  title        = {Few-Shot Parameter-Efficient Fine-Tuning is Better and Cheaper than
                  In-Context Learning},
  booktitle    = {NeurIPS},
  year         = {2022}
}
\bibliographystyle{icml2026}

\newpage
\appendix
\onecolumn
\section{Dataset Details}
\label{app:datasets}

In this section, we provide detailed descriptions of the datasets utilized in OpenMAG. To ensure a standardized evaluation, we explicitly detail the data source, the graph construction method (including node/edge definitions and preprocessing), and the specific composition of multimodal attributes (textual and visual features) for each dataset. We present the specific details and characteristics of each dataset below.

\textbf{Bili\_cartoon}~\cite{ninerec} is curated from the animated content channel of the Bilibili platform. The graph topology consists strictly of cartoon video nodes connected by undirected co-viewing edges, which are established when two videos are consumed by the same user within a temporally localized session. To ensure structural validity, we filtered out videos with fewer than five interactions. Textual features are encoded from video titles and uploader-generated tags using semantic encoders, while visual features are extracted from video cover thumbnails to capture the distinct artistic style of the animations. This dataset is primarily utilized for Link Prediction tasks.

\textbf{Bili\_dance}~\cite{ninerec} originates from the dance section of Bilibili. Nodes represent dance performance or tutorial videos. Co-viewing edges capture the trend similarity or sequential learning patterns of viewers (e.g., users watching consecutive tutorials). Textual attributes are derived from dense video descriptions and hashtags, while visual attributes are obtained from keyframes of the dance movements to encode choreographic dynamics. This dataset is primarily utilized for Link Prediction tasks.

\textbf{Bili\_food}~\cite{ninerec} is sourced from Bilibili's culinary community. The graph comprises food-related video nodes. Edges denote that two videos (e.g., recipes or reviews) were watched by the same user, reflecting dietary preferences or cooking interests. Textual features utilize recipe titles and ingredients lists, providing semantic context, while visual features focus on the presentation of dishes extracted from video covers. This dataset is primarily utilized for Link Prediction tasks.

\textbf{Bili\_movie}~\cite{ninerec} comes from the film and television section of Bilibili. Nodes denote movie clips or trailers, linked by audience co-consumption behaviors. We enrich the textual attributes with plot summaries and cast lists to capture narrative similarity, while visual attributes are extracted from official movie posters, providing rich semantic context for genre understanding. This dataset is primarily utilized for Link Prediction tasks.

\textbf{Bili\_music}~\cite{ninerec} is collected from the music video ecosystem on Bilibili. Nodes represent music videos (MVs) or live performances. Edges indicate shared listening habits among users. Textual features are encoded from song lyrics and artist descriptions, while visual features are derived from the MV content or album art, capturing the aesthetic atmosphere. This dataset is primarily utilized for Link Prediction tasks.

\textbf{DY}~\cite{ninerec} is sourced from TikTok (Douyin), a leading short-video platform. The graph contains short-video nodes linked by co-interaction patterns (e.g., liked by the same user). Given the multimodal nature of short videos, textual features are constructed from user captions and hashtags, while visual features are extracted from video frames, capturing the fast-paced visual dynamics typical of the platform. Preprocessing involved filtering low-frequency items to maintain graph connectivity. This dataset is primarily utilized for Link Prediction tasks.

\textbf{KU}~\cite{ninerec} originates from Kuaishou, another major short-video service. Nodes represent user-uploaded videos, with edges representing co-viewing relationships. Textual attributes are derived from the localized titles and comments, and visual attributes are obtained from raw video streams, reflecting distinct community content styles that differ from TikTok. This dataset is primarily utilized for Link Prediction tasks.

\textbf{QB}~\cite{ninerec} is derived from a specific online video service included in the NineRec collection. The graph consists of video items, where edges denote that items were consumed in the same user session. Textual features utilize video metadata and category tags, while visual features are extracted from thumbnails, serving as high-dimensional content descriptors. This dataset is primarily utilized for Link Prediction tasks.

\textbf{TN}~\cite{ninerec} originates from a widely-used news and content platform. Nodes represent video news or articles, connected by co-reading/viewing behaviors. Textual features are encoded from headlines, and visual features are derived from the accompanying cover images, forming a dense graph of topic-related content. This dataset is primarily utilized for Link Prediction tasks.

\textbf{Grocery}~\cite{Amazon2018} is sourced from Amazon's Grocery and Gourmet Food category. Nodes are food and household products, and edges indicate complementary purchasing habits derived from "also-bought" lists. Textual attributes are encoded from product titles and nutritional descriptions, while visual attributes are extracted from packaging images. This dataset is primarily utilized for Node Classification, where labels correspond to fine-grained product sub-categories (e.g., Beverages, Snacks).

\textbf{Toys}~\cite{Amazon2018} originates from Amazon's Toys and Games category. The graph connects toy product nodes via co-purchasing edges. Textual features are derived from product specifications and age recommendations, and visual features are obtained from product photos to identify visual variants. This dataset is primarily utilized for Node Classification, where labels represent specific toy types or game genres.

\begin{table*}[t]
\centering
\caption{OpenMAG Dataset Statistics}
\label{tab:app_dataset}

\fontsize{9pt}{10.5pt}\selectfont
\renewcommand{\arraystretch}{1.1}
\setlength{\tabcolsep}{10pt}

\begin{tabular*}{\textwidth}{
@{\extracolsep{\fill}}
c
c
c
c
c
c
c
}
\toprule
\textbf{Dataset} &
\textbf{Nodes} &
\textbf{Edges} &
\textbf{Labels} &
\textbf{Modalities} &
\textbf{Domain} &
\textbf{Train/Val/Test} \\
\midrule

Grocery       & 17,074 & 171,340 & 20 & Text,Visual & E-Commerce           & 60\%/20\%/20\% \\
RedditS       & 15,894 & 566,160 & 20 & Text,Visual & Social Media         & 60\%/20\%/20\% \\
Movies        & 16,672 & 218,390 & 20 & Text,Visual & E-Commerce              & 60\%/20\%/20\% \\
Toys          & 20,695 & 126,886 & 18 & Text,Visual & E-Commerce           & 60\%/20\%/20\% \\
Ele-fashion   & 97,766 & 199,602 & 12 & Text,Visual & E-Commerce           & 60\%/20\%/20\% \\
Goodreads-nc  & 685,294 & 7,235,048 & 11 & Text,Visual & Book Recommendation & 10\%/5\%/85\% \\
Cloth         & 125,839& 951,271 & - & Text,Visual & E-Commerce           & 60\%/20\%/20\% \\
Sports        & 50,250 & 356,202 & - & Text,Visual & E-Commerce           & 60\%/20\%/20\% \\
Bili\_cartoon & 4,724 & 18,660 & - & Text,Visual & Video Recommendation & 60\%/20\%/20\% \\
Bili\_dance   & 2,307 & 9,127 & - & Text,Visual & Video Recommendation & 60\%/20\%/20\% \\
Bili\_food    & 1,579  & 6,544 & - & Text,Visual & Video Recommendation & 60\%/20\%/20\% \\
Bili\_movie   & 3,509 & 11,618 & - & Text,Visual & Video Recommendation & 60\%/20\%/20\% \\
Bili\_music   & 6,038 & 21,592 & - & Text,Visual & Video Recommendation & 60\%/20\%/20\% \\
DY            & 8,299 & 35,627 & - & Text,Visual & Video Recommendation & 60\%/20\%/20\% \\
KU            & 5,370  & 22,052 & - & Text,Visual & Video Recommendation & 60\%/20\%/20\% \\
QB            & 6,121 & 24,145 & - & Text,Visual & Video Recommendation & 60\%/20\%/20\% \\
TN            & 3,334 & 13,344 & - & Text,Visual & Video Recommendation & 60\%/20\%/20\% \\
SemArt        & 21,382 & 1,216,432 & - & Text,Visual & Art Networks     & 60\%/20\%/20\% \\
Flickr30k     & 31,783 & 181,151 & - & Text,Visual & Image Networks      & 60\%/20\%/20\% \\

\bottomrule
\end{tabular*}
\end{table*}

\textbf{Cloth}~\cite{Amazonreview1,Amazonreview2} comes from Amazon's Clothing, Shoes, and Jewelry category. This large-scale graph links fashion items based on visual compatibility or co-purchase history. Textual features include brand names and fabric details, while visual features are extracted from high-resolution model images. This dataset is primarily utilized for Node Classification, where labels indicate fashion categories or styles.

\textbf{Sports}~\cite{Amazonreview1,Amazonreview2} is sourced from Amazon's Sports and Outdoors category. Nodes represent athletic gear and equipment, and edges denote functional complementarity. Textual attributes utilize technical specifications, and visual attributes are derived from product images to capture design and utility. This dataset is primarily utilized for Link Prediction.

\textbf{Ele-fashion}~\cite{Amazonreview1,Amazonreview2} is a heterogeneous graph constructed by merging Amazon's Electronics and Fashion categories. Nodes represent items from both domains, connected by cross-category co-purchasing links. Textual features combine technical specs and style descriptions, while visual features come from product images. This dataset is primarily utilized for Node Classification, where labels correspond to high-level domain categories.

\textbf{RedditS}~\cite{RedditS} is collected from the Reddit social media platform. The graph consists of post nodes, where edges represent threading relationships (e.g., posts within the same discussion). Textual features are encoded from post titles and body content, while visual features are extracted from images embedded in the posts. This dataset is primarily utilized for Node Classification, where labels represent the subreddit community to which a post belongs.

\textbf{Movies}~\cite{Amazon2018} is collected from Amazon's Movies and TV category. Nodes represent DVD or Blu-ray products, with edges reflecting co-purchasing behavior. Textual attributes include plot synopses and customer reviews, while visual attributes are derived from official cover art. This dataset is primarily utilized for Node Classification, where labels denote movie genres (e.g., Action, Drama).

\textbf{SemArt}~\cite{semart} is sourced from the Web Gallery of Art, a large-scale collection of fine-art paintings. Nodes represent artworks, and edges are constructed based on shared attributes (e.g., same artist, time period, or school). Textual features are encoded from the artwork titles and historical comments, while visual features are extracted from the digital images of the paintings. This dataset serves as a benchmark for G2Image tasks, challenging models to generate or retrieve artistic images based on graph-structured attributes.

\textbf{Flickr30k}~\cite{flickr30k} is a canonical dataset for image-text reasoning. In OpenMAG, we construct a graph where nodes represent image regions and caption phrases, linked by semantic grounding annotations. Edges reflect the alignment between visual objects and textual descriptions. This dataset allows for the evaluation of generative capabilities and is primarily used for G2Text tasks, specifically image captioning and region description generation.

\textbf{Goodreads-nc}~\cite{books-nc1,books-nc2} originates from the Goodreads book review website. The graph is constructed with book nodes, where edges represent "co-shelving" relationships (books placed on the same user shelf). Textual attributes are derived from book summaries and editorial reviews, while visual attributes are extracted from book covers. This dataset is primarily utilized for Node Classification, where labels correspond to literary genres.

\section{Modality Encoders}
\label{app:feature-encoders}

The OpenMAG benchmark integrates a diverse set of feature extractors to extract and construct essential feature representations. This collection includes large-scale pretrained encoders operating in frozen mode to leverage their generalizable knowledge, alongside compact and trainable models that provide adaptable feature extraction for specific scenarios. Collectively, these extractors form the foundational encoding layer that supplies the essential visual, textual, and cross-modal features for all subsequent multimodal reasoning and fusion processes within OpenMAG.

\subsection{Visual Encoders}
This category encompasses models dedicated to extracting hierarchical and semantic features from visual inputs. They are further divided into frozen encoders that offer stable, pre-trained representations and trainable encoders that allow for task-specific adaptation and architectural exploration. Notably, in addition to dedicated graph-based visual models, we also employ fine-tuned ViT as trainable backbones to ensure visual representations are optimized for the target domains.

\textbf{ConvNeXtV2-Base-22k-224}~\cite{convnextv2} is a modern convolutional neural network that evolves the ResNet architecture with design principles from Vision Transformers. The V2 version incorporates techniques like Global Response Normalization (GRN) to enhance feature diversity and representation power. Pre-trained on ImageNet-22K, this frozen encoder provides efficient and powerful hierarchical visual features.

\textbf{SwinV2-Large-Patch4-Window12-192-22k}~\cite{swin} is a hierarchical Vision Transformer that uses shifted windows to efficiently model at various scales while allowing cross-window connections. The V2 version introduces techniques like log-spaced continuous position bias to better handle high-resolution images. Pre-trained on ImageNet-22K, this large frozen encoder is adept at capturing multi-scale visual patterns.

\textbf{DINOv2-Large}~\cite{dinov2} is a self-supervised visual encoder trained using the DINOv2 framework, which leverages knowledge distillation from a teacher network without the use of manually annotated labels. It produces high-performance, general-purpose visual features that are particularly strong in capturing geometric and semantic information. The frozen DINOv2-Large encoder provides robust 1024-dimensional visual representations.

\textbf{ViT-Base-Patch16-224}~\cite{vit} is a standard Vision Transformer encoder that splits an image into fixed-size patches and processes them with a standard Transformer encoder. The base variant with patch size 16 and input resolution 224x224 offers a good balance between performance and computational cost. This frozen encoder serves as a fundamental Transformer-based baseline for visual feature extraction. To further enhance adaptability, we also utilize ViT as a trainable backbone. By enabling gradient updates across the encoder layers, we optimize the patch embeddings and self-attention maps to learn features specific to the visual distributions of our graph datasets, bridging the gap between general pre-training and the specific visual semantics required by our tasks.

\textbf{ViG}~\cite{vig} is a novel architecture that models an image as a graph structure, where image patches are represented as nodes, and edges connect semantically or spatially related nodes. This departure from the grid-like inductive bias of CNNs or the sequential patch processing of Vision Transformers allows ViG to explicitly capture long-range dependencies and complex topological relationships within an image. It typically employs graph convolutional layers or transformers to perform message passing between nodes. In our framework, the ViG encoder is trainable, allowing the model to learn task-specific graph structures and feature representations directly from the data, offering a flexible and powerful alternative for visual feature extraction.


\textbf{PyramidTNT}~\cite{tnt} is a hierarchical vision model that integrates Transformer and CNN design philosophies, specifically tailored for handling multi-scale information in images. It typically extracts features at multiple levels (scales), with lower layers capturing local details and higher layers integrating global semantics. This pyramid structure, combined with a Transformer or CNN backbone, enables the model to effectively handle objects of varying sizes and complex visual scenes. In implementation, PyramidTNT may embed Transformer blocks within a Feature Pyramid Network or introduce multi-stage downsampling akin to CNNs into a Transformer. In our experiments, this trainable encoder allows the network to adaptively learn and fuse multi-scale visual features ranging from pixel-level details to scene-level semantics, which is crucial for tasks requiring fine-grained understanding.

\subsection{Text Encoders}

This category includes encoders specialized in processing and representing textual information. They convert raw text into dense, contextualized feature vectors that capture linguistic semantics and structure, serving as the primary language understanding component. Crucially, beyond utilizing frozen extractors, we specifically fine-tune BERT-based architectures as trainable backbones to enable domain-adaptive text encoding directly from the benchmark data.

\textbf{XLM-RoBERTa-Base}~\cite{roberta} is a multilingual variant of the RoBERTa model, pre-trained on a massive corpus of text in 100 languages using the masked language modeling objective. It is designed to generate high-quality contextual representations that are transferable across languages without the need for parallel data. This cross-lingual capability makes it a robust baseline for various multilingual NLP tasks, as evidenced by its application in sequence labeling and named entity recognition in international evaluations like SemEval. With an output dimension of 768, the frozen XLM-RoBERTa encoder in our setup provides language-agnostic textual features, facilitating consistent processing of multilingual inputs within the multimodal pipeline.

\textbf{Llama-3.2-1B-Instruct}~\cite{llama3.2} is a compact, instruction-tuned language model from Meta's Llama 3.2 series. With 1 billion parameters, it is designed for efficiency while maintaining strong conversational and instruction-following capabilities. The model is optimized for tasks requiring precise adherence to user prompts and generating helpful, safe responses. Despite its smaller size, it benefits from the scalable Transformer architecture and extensive pre-training data characteristic of the Llama family. In this work, its parameters are frozen. The encoder produces a comparatively high-dimensional output of 2048, offering dense textual representations suitable for fusion with features from other modalities in resource-constrained scenarios.

\textbf{Facebook OPT-125M}~\cite{opt} is a small-scale autoregressive language model developed by Meta with 125 million parameters. It is part of the OPT suite designed to democratize access to large language models by providing open-access pre-trained models of various sizes. OPT-125M follows the standard decoder-only Transformer architecture and is pre-trained on a diverse blend of publicly available text datasets. While its capacity is limited compared to larger models, it provides an efficient and accessible baseline for text understanding and generation. The frozen OPT-125M encoder in our framework provides 768-dimensional textual features, serving as a lightweight linguistic component.

\textbf{T5-Large}~\cite{t5} is an encoder-decoder model pre-trained on a diverse set of tasks formulated as a text-to-text problem. Its unified framework converts every language problem (translation, summarization, question answering, etc.) into feeding text as input and generating target text as output. The “Large” variant contains hundreds of millions of parameters. The encoder component of T5 produces dense representations that are highly effective for transfer learning. In our approach, we utilize the frozen T5-Large encoder to process input text, capturing complex semantic and syntactic information to facilitate deep integration with other modalities.

\textbf{bert-base-nli-mean-tokens}~\cite{sbert} is a sentence-transformers model fine-tuned on Natural Language Inference (NLI) data. It is specifically optimized to produce high-quality sentence embeddings by using mean pooling over the output tokens. This frozen encoder excels at capturing semantic textual similarity, making it particularly effective for tasks requiring sentence-level understanding and comparison within our multimodal pipeline. In addition to its use as a frozen extractor, we also deploy it as a trainable backbone within our framework. By performing end-to-end fine-tuning on the transformer layers, we allow the model to adaptively adjust its attention mechanisms to capture domain-specific linguistic nuances and task-relevant semantics that static pre-trained weights might overlook, offering a flexible alternative to fixed embeddings.

\subsection{Multimodal Encoders}
This category comprises encoders that are inherently designed to process and align information from both visual and textual modalities. They provide joint representations that are foundational for cross-modal understanding and reasoning within the OpenMAG framework.

\textbf{Qwen2-7B-Instruct}~\cite{qwenvl} is a large-scale, instruction-tuned multimodal language model from the Qwen2 series. It is based on the Transformer architecture and incorporates key improvements including the SwiGLU activation function, attention QKV bias, and Grouped Query Attention (GQA). The model supports an extensive context length of up to 131,072 tokens. Compared to its predecessor and other contemporary open-source models, Qwen2-7B-Instruct demonstrates competitive or superior performance across a wide range of benchmarks, excelling in language understanding, generation, coding (e.g., 79.9 on HumanEval), mathematics, and multilingual tasks. In this work, its parameters are frozen to leverage its robust, pre-trained representational capabilities while providing a 3584-dimensional feature vector for downstream fusion tasks.

\textbf{Qwen2.5-3B-Instruct}~\cite{qwenvl} is a compact yet powerful model from the latest Qwen2.5 series, featuring approximately 3.09 billion parameters. It represents a significant enhancement over Qwen2, with notable improvements in knowledge capacity, coding, and mathematical reasoning, attributed to training with specialized expert models in these domains. The model excels in instruction following, long-context processing (supporting up to 128K tokens), and understanding/generating structured data (e.g., JSON). Its architecture employs RoPE, RMSNorm, and a GQA configuration with 16 query heads and 2 key-value heads. With an output dimension of 2048, this frozen encoder provides a balance between model capacity and feature richness for multimodal integration.

\textbf{CLIP-ViT-Large-Patch14}~\cite{clip} is a seminal vision-language model that learns visual and textual representations in a shared embedding space through contrastive pre-training on large-scale image-text pairs. The visual encoder is based on a Vision Transformer (ViT) with a large backbone and a patch size of 14x14. This architecture enables the model to capture rich, contextualized visual features that are semantically aligned with text. In image classification tasks, such as on ImageNet-1K, models based on this encoder have achieved top-1 accuracy exceeding 88\%. In our framework, this encoder remains frozen, providing robust 768-dimensional visual features that are inherently aligned with linguistic concepts, serving as a cornerstone for cross-modal understanding.

\textbf{Llama-3.2-11B-Vision-Instruct}~\cite{llama3.2} is a large-scale vision-language instruction-following model. Based on the Llama 3.2 architecture, it is pre-trained and fine-tuned to understand and reason over both visual inputs and textual instructions. This frozen encoder provides powerful cross-modal representations for tasks requiring joint comprehension of images and text.

\textbf{ImageBind-Huge}~\cite{imagebind} is a groundbreaking multimodal encoder from Meta AI that learns a joint embedding space across six different modalities: image, text, audio, depth, thermal, and IMU data. By leveraging the natural pairing of images with other modalities during pre-training, it enables emergent zero-shot recognition across all modalities. The frozen ImageBind-Huge encoder in our framework provides a unified, semantically rich feature representation that links visual and non-visual concepts.

\section{Model Framework}
\label{app:model_framework}

In this section, we provide detailed descriptions of the models integrated into OpenMAG. We first introduce the conventional GNN backbones that serve as unimodal baseline. Subsequently, we present the specialized library of MAG learning models, categorized by their core architectural paradigms.

\subsection{Conventional Graph Neural Networks}
\label{app:general_gnns}

This subsection details the foundational GNN architectures utilized within OpenMAG. These models primarily focus on topological representation learning and are employed either as standalone baselines to assess the necessity of multimodal integration or as structural backbones within more complex frameworks.

\noindent \textbf{GCN}~\cite{GCN}: GCN is a seminal model for semi-supervised learning on graphs. It operates via a layer-wise propagation rule that efficiently aggregates feature information from a node's immediate neighbors. The model simplifies spectral graph convolutions through a first-order approximation, making it scalable and effective for tasks like node classification in citation networks.

\noindent \textbf{GAT}~\cite{GAT}: GAT introduces an attention mechanism to graph neural networks. Instead of using fixed, uniform weighting for neighbors (like GCN), it assigns dynamic, learned importance scores to each neighbor's features. This allows the model to focus on more relevant parts of a node's neighborhood. The architecture supports multi-head attention for stability and is inherently inductive, meaning it can generalize to unseen graph structures.

\noindent \textbf{GraphSAGE}~\cite{GraphSAGE}: GraphSAGE is a framework for inductive representation learning on large graphs. Its core idea is to learn aggregator functions that can sample and combine features from a node's local neighborhood. Rather than training individual node embeddings, it trains a function that generates embeddings by aggregating neighbor information. This enables it to efficiently produce embeddings for entirely new, unseen nodes, making it highly suitable for dynamic, evolving graphs.

\noindent \textbf{RevGAT}~\cite{RevGAT}: RevGAT is designed to train extremely deep graph attention networks by incorporating reversible connections. This architectural innovation dramatically reduces memory consumption during training, as it allows activations to be recomputed during the backward pass rather than stored. This enables the training of GNNs with hundreds or even thousands of layers, which was previously infeasible, leading to more powerful and overparameterized models.

\noindent \textbf{GIN}~\cite{GIN}: GIN is a theoretically motivated model designed to maximize the expressive power of GNNs. Its architecture is proven to be as powerful as the Weisfeiler-Lehman graph isomorphism test in distinguishing graph structures. It achieves this by using a simple injective multiset function for neighbor aggregation, typically implemented with a Multi-Layer Perceptron (MLP). It is particularly effective for graph classification tasks.

\noindent \textbf{ChebNet}~\cite{ChebNet}: ChebNet is an early and influential spectral-based graph convolutional network. It leverages Chebyshev polynomials to approximate localized spectral filters on the graph Laplacian. This approach avoids the need for computationally expensive eigendecomposition and allows the filters to be spatially localized, meaning they operate within a \(K\)-hop neighborhood of a node. It laid important groundwork for efficient spectral graph convolutions.

\noindent \textbf{GCNII}~\cite{gcn2}: GCNII addresses the common problem of over-smoothing in deep GCNs, where node representations become indistinguishable after many layers. It extends the vanilla GCN with two key techniques: Initial Residual connections (linking to the initial input features) and Identity Mapping (incorporating an identity matrix in the weight transformation). These techniques help preserve individual node information and allow for the successful training of much deeper GCN models.

\noindent \textbf{GATv2}~\cite{gat2}: GATv2 is an improved version of the original GAT that fixes a key expressive limitation. The authors show that GAT's attention mechanism is static—the ranking of attention scores for a node's neighbors does not truly depend on the node itself. GATv2 modifies the internal order of operations to create a dynamic attention mechanism. This makes the model strictly more expressive, allowing it to learn more complex relationships where the importance of a neighbor varies depending on the central query node.

\subsection{Advanced MAG Learning Models}
\label{app:mag_models}

To provide a systematic understanding of the diverse methodologies in Multimodal Attributed Graph (MAG) learning, we categorize the integrated models in OpenMAG into a unified taxonomy comprising three distinct paradigms based on their core architectural mechanisms.
\textbf{Graph-enhanced Models} (e.g., GraphMAE2, DMGC, DGF, MIG-GT) prioritize the refinement of topological representation learning. These methods typically employ advanced spectral filtering, self-supervised masking, or structure reconstruction objectives to capture robust structural dependencies and mitigate modality-induced noise.
\textbf{Multimodal-enhanced Models} (e.g., MMGCN, MGAT, GSMN, MGNet, MHGAT, LGMRec) focus on the explicit synergy of heterogeneous modalities. These approaches incorporate specialized cross-modal attention, gating mechanisms, or co-attention layers to model the intricate interplay between node attributes and graph topology.
\textbf{MLLM-based Models} (e.g., UniGraph2, NTSFormer, MLaGA, GraphGPT-O, Graph4MM, InstructG2I) represent the emerging paradigm that leverages Multimodal Large Language Models (MLLMs) or Transformers as powerful backbones. These frameworks align graph topology with high-dimensional semantic spaces by treating structural information as auxiliary tokens, employing strategies such as neighborhood linearization or soft-token projection.
Below, we provide detailed descriptions of these models.

\noindent \textbf{GraphMAE2}~\cite{GraphMAE2}: GraphMAE2 is a decoding-enhanced masked self-supervised graph learner that improves upon GraphMAE. It addresses the vulnerability of masked feature reconstruction to less discriminative or noisy node features. The framework introduces two key strategies: multi-view random re-mask decoding, which introduces randomness into the reconstruction process to reduce overfitting to input features, and latent representation prediction, which enforces reconstruction in the embedding space for more informative targets. These strategies act as regularizations, making GraphMAE2 consistently outperform prior self-supervised methods across various graph scales.

\noindent \textbf{DMGC}~\cite{DMGC}: DMGC is a framework for Disentangled Multimodal Graph Clustering. It addresses the challenge of hybrid neighborhood patterns in real-world multimodal graphs, where nodes exhibit both homophilic (same-class) and heterophilic (different-class) connections. DMGC disentangles the original graph into a cross-modality homophily-enhanced graph and modality-specific heterophily-aware graphs. It then employs a multimodal dual-frequency fusion mechanism that jointly filters these graphs through low-pass and high-pass operations to capture intra-class commonalities and inter-class distinctions. A suite of self-supervised alignment objectives guides the learning for effective clustering without requiring labels.

\noindent \textbf{MIG-GT}~\cite{MIG-GT}: MIG-GT proposes Modality-Independent Graph Neural Networks with Global Transformers for multimodal recommendation. It observes that optimal receptive fields for GNNs can vary across modalities. Thus, it employs separate GNNs with independent hop numbers for each modality. To complement the potentially limited receptive fields of these GNNs, it introduces a Sampling-based Global Transformer that integrates global information via uniform sampling. This combination of modality-specific local aggregation and globally-aware modeling enhances the performance of multimodal recommendation systems.

\noindent \textbf{DGF}~\cite{DGF}: DGF is a framework for Cross-Contrastive Clustering on Multimodal-Attributed Graphs. It tackles the limitations of traditional multi-view clustering methods when dealing with features from large pre-trained models, which may have low inter-modality correlation and high noise. The core innovation is the Dual Graph Filtering scheme, which incorporates a feature-wise denoising component into node representation learning. Combined with a tri-cross contrastive training strategy across modalities, neighborhoods, and communities, DGF learns robust and discriminative representations, significantly outperforming baselines on clustering tasks.

\noindent \textbf{MMGCN}~\cite{MMGCN}: MMGCN (Multi-modal Graph Convolution Network) is a pioneering framework designed for personalized micro-video recommendation by explicitly modeling user preferences across different content modalities (e.g., visual, acoustic, textual). Unlike prior multimedia recommendation methods that treat multi-modal features as a unified whole, MMGCN constructs separate user-item bipartite graphs for each modality. Within each modality-specific graph, it applies graph convolution operations to propagate and aggregate information from interacted items to users, thereby learning distinct modal-specific user representations. These representations are then combined through a novel fusion layer that integrates structural information, intrinsic modal features, and a shared user ID embedding. This design enables MMGCN to capture fine-grained, modality-aware user preferences and leverage high-order connectivity in the interaction graphs. Extensive experiments on real-world micro-video datasets demonstrate that MMGCN significantly outperforms state-of-the-art recommendation baselines, validating its effectiveness in harnessing multi-modal signals for personalized recommendation.

\noindent \textbf{MGAT}~\cite{MGAT}: MGAT is a Multimodal Graph Attention Network designed for recommendation. It addresses the limitation of prior models that treat information from all neighbors equally within parallel multimodal interaction graphs. MGAT incorporates a gated attention mechanism to adaptively identify varying importance scores of different modalities (e.g., visual, acoustic, textual) to a user's preference. This allows the model to disentangle personal interests at a granular level, capture complex interaction patterns, and reduce the influence of noisy information, leading to more robust and accurate recommendations.

\noindent \textbf{GSMN}~\cite{GSMN}: GSMN (Graph Structured Matching Network) is a model for fine-grained image-text matching. It explicitly structures an image by modeling objects, attributes, and their relations as a graph. The core innovation involves performing node-level matching to associate elements (objects, relations, attributes) across modalities, followed by structure-level matching that fuses these neighborhood associations to infer the overall correspondence. This structured approach enables learning beyond coarse object co-occurrence, significantly improving fine-grained phrase-to-region matching accuracy.

\noindent \textbf{MHGAT}~\cite{MHGAT}: MHGAT (Multimodal Heterogeneous Graph Attention Network) is a model designed to learn representations from multimodal heterogeneous graphs, which contain various node/edge types and features from different modalities (e.g., text and images). To adaptively capture heterogeneous structural information without relying on manually defined meta-paths, MHGAT groups neighboring nodes by edge type and performs intra-group aggregation (using mean or max pooling) to form edge-level embeddings. These embeddings are then transformed into a shared feature space and aggregated via an edge-level attention mechanism that learns the importance of different relation types. For multimodal fusion, a modality-level attention network dynamically assigns weights to different modalities (e.g., image vs. text) based on their relevance to the node. To capture higher-order neighborhood information while mitigating over-smoothing, the model incorporates residual connections and dense connections across layers, concatenating outputs from all layers as the final node representation. Evaluated on constructed datasets (IMDB, AMAZON, DOUBAN) for node classification, clustering, and visualization, MHGAT demonstrates superior performance compared to homogeneous and unimodal heterogeneous graph baselines.

\noindent \textbf{MGNet}~\cite{MGNet}: MGNet is a Multiplex Graph Convolutional Network for multimodal brain network analysis. The model integrates tensor representation to extract the latent common structures from a set of multimodal brain networks (e.g., from different imaging techniques). It then employs multiplex GCNs to generate modality-specific representations that capture the unique graph structures within each modality. This approach facilitates an intuitive analysis of the human connectome across modalities and has demonstrated state-of-the-art performance in classification tasks on neurological disorder datasets.

\noindent \textbf{LGMRec}~\cite{LGMRec}: LGMRec is a multimodal recommender that jointly models Local and Global user interests via graph learning. It addresses the coupling of collaborative and multimodal signals in shared user embeddings and the sparsity problem in local interest modeling. The model uses a local graph embedding module to independently learn collaborative and modality-related embeddings. Additionally, a global hypergraph embedding module captures insightful global dependency relations among all users and items. The combination of these decoupled local and robust global embeddings enhances recommendation accuracy and robustness.

\noindent \textbf{UniGraph2}~\cite{UniGraph2}: UniGraph2 is a novel cross-domain graph foundation model designed for MAGs. It introduces a unified embedding space that effectively integrates diverse node modalities (e.g., text, images) and the underlying graph structure. The framework employs frozen modality-specific encoders to process raw data, followed by a Mixture of Experts (MoE) module for cross-domain and cross-modality feature alignment. A GNN then aggregates these aligned features within the graph context. Pre-trained using a multimodal masked prediction task alongside structural reconstruction objectives, UniGraph2 learns general and transferable representations. It demonstrates strong performance and generalization across various downstream tasks, including representation learning, few-shot transfer, and multimodal generation, without requiring task-specific retraining.

\noindent \textbf{GraphGPT-O}~\cite{GraphGPT-O}: GraphGPT-O is a multimodal large language model designed for synergistic comprehension and generation on Multimodal-Attributed Graphs. It addresses key challenges in applying MLLMs to MAGs, including graph size explosion, non-Euclidean graph nature, hierarchical modality dependency, and inference dependency. The model incorporates a Personalized PageRank-based sampling strategy to extract relevant neighbors, and a hierarchical aligner with node-level and graph-level Q-Formers to capture deep graph structure and multimodal dependencies. Built upon DreamLLM, GraphGPT-O supports dual inference strategies (sequential and parallel) for joint text-image generation and demonstrates superior performance on tasks such as artwork and product generation across multiple benchmark datasets.

\noindent \textbf{MLaGA}~\cite{MLaGA}: MLaGA (Multimodal Large Language and Graph Assistant) is a novel framework that extends Large Language Models (LLMs) to reason over multimodal graphs, where nodes possess both textual and visual attributes. It introduces a two-stage approach: a structure-aware multimodal aligner that unifies image and text features into a shared latent space via contrastive graph pre-training, followed by multimodal graph instruction tuning, which integrates these aligned representations and graph structure into the LLM through lightweight projectors and task-aware demonstration templates. This design enables MLaGA to effectively handle the heterogeneity of multimodal node attributes while leveraging graph connectivity, achieving strong performance and generalization across diverse graph learning tasks.

\noindent \textbf{Graph4MM}~\cite{Graph4MM}: Graph4MM is a framework for weaving multimodal learning with structural information. It moves beyond simple one-to-one image-text pair modeling by representing complex many-to-many relationships within a multimodal attributed graph. Its core innovations are Hop-Diffused Attention, which integrates multi-hop structural information into self-attention via causal masking and a diffusion mechanism to avoid over-smoothing, and MM-QFormer, a multi-mapping querying transformer for principled cross-modal fusion. The framework demonstrates that using graph structure to guide intra- and inter-modal interactions is more effective than treating the graph as a standalone modality, leading to superior performance in both generative and discriminative tasks.

\noindent \textbf{NTSFormer}~\cite{NTSFormer}: NTSFormer (Neighbor-to-Self Graph Transformer) is a self-teaching Graph Transformer framework designed for multimodal isolated cold-start node classification, where test nodes are both isolated (no edges) and may have missing modalities. To address this dual challenge, NTSFormer introduces a novel self-teaching paradigm within a unified Transformer architecture. It employs a cold-start attention mask to produce two predictions per node: a student prediction based solely on the node's own multimodal features (self-information), and a teacher prediction that additionally incorporates neighbor information. This design allows the model to supervise itself during training without degrading to a simpler MLP, thereby retaining Transformer capacity. To handle diverse graph data and missing modalities, NTSFormer performs a one-time multimodal graph pre-computation to convert structural and feature information into fixed-length token sequences, which are then processed by a Mixture-of-Experts (MoE) input projection layer for effective fusion. Experimental results on multimodal graph benchmarks demonstrate that NTSFormer consistently outperforms existing GNNs, Graph Transformers, and teacher-student distillation methods, showcasing its robustness and generalization ability in challenging cold-start scenarios.

\noindent \textbf{InstructG2I}~\cite{instructg2i}: InstructG2I is a graph context-conditioned diffusion model designed for synthesizing images from Multimodal-Attributed Graphs. It tackles key challenges such as graph size explosion, entity dependencies, and controllability. The framework first employs a semantic Personalized PageRank-based neighbor sampling method to select informative neighboring nodes by combining structural and semantic relevance. A Graph-QFormer encoder then adaptively encodes these neighbors into a fixed set of graph prompt tokens, which condition the denoising process of a Stable Diffusion backbone alongside text prompts. Additionally, the model introduces graph classifier-free guidance, allowing fine-grained control over the strength of graph conditioning and the blending of multiple graph influences (e.g., artistic styles). Evaluated on art, e-commerce, and literature graphs, InstructG2I demonstrates superior image generation quality and controllability compared to existing text-to-image and image-to-image baselines.

\section{Downstream Tasks}
\label{app:downstream_tasks}

The OpenMAG benchmark is designed to comprehensively evaluate the capabilities of multimodal graph learning models across a spectrum of downstream tasks. These tasks are categorized into \textbf{Graph-Based Tasks}, which emphasize topological reasoning and node-level representation learning, and \textbf{Modality-Based Tasks}, which focus on cross-modal interaction, alignment, and generation. Each task is associated with specific evaluation metrics, which are formally defined in Sec.~\ref{app:evaluation_strategies}. The implementation details and hyperparameter settings are provided in Sec.~\ref{app:task_hyperparameters}.

\subsection{Graph-Based Tasks}

\textbf{Node Classification} is a fundamental supervised learning task that evaluates the model's ability to learn discriminative node representations. Given a multimodal graph with node attributes and edge structures, the model encodes each node into a low-dimensional embedding. These embeddings are passed through a projection head, typically a Multi-Layer Perceptron (MLP) or a linear classifier, followed by a Softmax layer to produce class probabilities. The model is trained by minimizing the discrepancy between predicted labels and ground-truth annotations. This task assesses how effectively a model can suppress irrelevant modality noise while capturing the underlying semantic structure of the graph. Performance is evaluated using \textbf{Accuracy} and \textbf{F1-Score} (see Sec.~\ref{app:evaluation_strategies}).

\textbf{Link Prediction} is a self-supervised or semi-supervised task that measures the model’s ability to infer missing or potential edges in a graph. The model computes similarity scores between pairs of node embeddings, using functions such as dot product or cosine similarity, and assigns higher scores to true edges than to negative samples. In multimodal graphs, this task requires the alignment of structural proximity with semantic similarity across modalities. The task is evaluated using ranking-based metrics, including \textbf{Mean Reciprocal Rank (MRR)} and \textbf{Hits@K} (see Sec.~\ref{app:evaluation_strategies}).

\textbf{Node Clustering} evaluates the quality of learned node representations in an unsupervised setting, following the protocol in DMGC~\cite{DMGC}. This task requires the model to partition nodes into disjoint semantic groups without label supervision. The process involves disentangling the multimodal graph into homophilic and heterophilic views to capture hybrid neighborhood patterns, followed by a dual-frequency fusion mechanism that integrates high-pass and low-pass filtered signals. The model is optimized using a joint objective function comprising reconstruction, contrastive alignment, and a specific clustering loss (e.g., minimizing the Kullback-Leibler divergence between soft cluster assignments and a target distribution). This task assesses the model's capability to uncover latent community structures and resolve complex topological interactions. Performance is evaluated using \textbf{Accuracy}, \textbf{Normalized Mutual Information (NMI)}, and \textbf{Adjusted Rand Index (ARI)} (see Sec.~\ref{app:evaluation_strategies}). 

\subsection{Modality-Based Tasks}

\textbf{Modality Matching} is a supervised task that determines whether a pair of inputs from different modalities, such as an image and a text description associated with a specific node, are semantically correlated. The process begins by projecting heterogeneous features into a unified embedding space using modality-specific encoders. A matching score is then computed for each pair, typically via a dot product or a non-linear classifier, and the model is trained to distinguish positive ground-truth pairs from randomly sampled negative pairs (e.g., mismatched image-text combinations). This task serves as a fundamental validation of the model's ability to verify instance-level cross-modal correspondence. Performance is evaluated using \textbf{Accuracy} and \textbf{Mean Reciprocal Rank (MRR)} (see Sec.~\ref{app:evaluation_strategies}).

\textbf{Modality Retrieval} evaluates the model's ability to search for relevant instances across image and text modalities in a ranking scenario. Specifically within OpenMAG, this task focuses on Image-to-Text and Text-to-Image retrieval. Given a query from one modality (e.g., an image), the model encodes both the query and all candidate instances from the other modality (e.g., texts) into a shared latent space. It then computes pairwise similarity scores and ranks the candidates in descending order of relevance. This task assesses the robustness of the learned representations for cross-modal information retrieval. Performance is measured using \textbf{MRR} and \textbf{Hits@K} (see Sec.~\ref{app:evaluation_strategies}).

\textbf{Modality Alignment} assesses the precise, fine-grained geometric consistency between feature distributions of distinct modalities. Unlike global distribution matching, this task requires the model to align detailed semantic features—such as mapping specific visual regions to corresponding textual phrases—within the shared embedding manifold. The procedure involves optimizing an alignment objective that minimizes the discrepancy between paired modality representations at a granular level, ensuring that subtle semantic distinctions are preserved across modalities. This task reflects the effectiveness of cross-modal knowledge transfer and semantic coherence. It is quantitatively evaluated using the \textbf{CLIP-Score} (see Sec.~\ref{app:evaluation_strategies}).

\textbf{Graph-to-Text (G2Text)} is a generative task that requires the model to produce natural language descriptions conditioned on graph-structured multimodal inputs. Unlike traditional multimodal tasks that assume simple one-to-one mappings, this task addresses complex \emph{many-to-many relationships} where a target node interacts with multiple multimodal neighbors. The process follows the \textbf{MMGL} framework~\cite{MMGL}: First, multimodal neighbor information is processed via neighbor encoding, utilizing frozen encoders to map diverse modalities into a compatible embedding space. Second, the topological context is captured through graph structure encoding, employing techniques such as GNNs or Laplacian position encodings(LPE) to represent structural dependencies. Finally, these structure-aware multimodal signals are infused into a pretrained LLM using parameter-efficient fine-tuning methods (e.g., LoRA or Prefix Tuning). This task evaluates whether structured graph information can be faithfully translated into coherent and semantically accurate text. Performance is assessed using standard \textbf{NLG metrics}, including \textbf{BLEU-4}, \textbf{ROUGE-L}, and \textbf{CIDEr} (see Sec.~\ref{app:evaluation_strategies}).

\textbf{Graph-to-Image (G2Image)} focuses on synthesizing images conditioned on Multimodal-Attributed Graphs, addressing the challenge of generating visual content that reflects both textual descriptions and complex graph associations. Following the \textbf{InstructG2I} framework~\cite{instructg2i}, the workflow proceeds in three stages. First, to handle combinatorial graph complexity, the model employs Semantic PPR-based Neighbor Sampling, which selects the most informative neighboring nodes by combining structural relevance (via Personalized PageRank) and semantic similarity. Second, a Graph-QFormer encoder transforms these sampled neighbors into graph conditioning tokens, utilizing self-attention to capture image-image dependencies and cross-attention for text-image alignment. Finally, these tokens guide a latent diffusion model (e.g., Stable Diffusion) through a Graph Classifier-Free Guidance mechanism, which dynamically balances the influence of textual prompts and graph contexts during the denoising process. This task evaluates the model's ability to generate images that are consistent with specific styles or semantic categories defined by the graph structure. Performance is quantitatively assessed using \textbf{CLIP-Score} and \textbf{DINOv2-Score} for instance-level consistency to measure distribution-level quality (see Sec.~\ref{app:evaluation_strategies}).

\section{Evaluation Strategies}
\label{app:evaluation_strategies}

\textbf{Accuracy (Acc)} measures the proportion of correctly predicted samples over the entire evaluation set. It calculates the ratio of true positives and true negatives to the total number of samples. Let $N$ denote the total number of samples, $y_i$ the ground-truth label of the $i$-th sample, $\hat{y}_i$ the corresponding predicted label, and $\mathbb{I}(\cdot)$ the indicator function. This metric serves as the primary benchmark for \textbf{Node Classification} and \textbf{Modality Matching} tasks, providing a direct and intuitive measure of model performance when the class distribution is relatively balanced.
\begin{equation}
    \text{Acc} = \frac{1}{N} \sum_{i=1}^{N} \mathbb{I}(\hat{y}_i = y_i).
\end{equation}

\textbf{F1-Score} is the harmonic mean of Precision and Recall, designed to provide a robust evaluation under class imbalance. Precision measures the fraction of correctly predicted positive samples among all predicted positives, while Recall measures the fraction of correctly predicted positives among all actual positives. This metric is particularly vital for \textbf{Node Classification} on real-world graphs where long-tailed class distributions are common, as it effectively balances the trade-off between false positives and false negatives.
\begin{equation}
    \text{F1} = 2 \cdot \frac{\text{Precision} \cdot \text{Recall}}{\text{Precision} + \text{Recall}}.
\end{equation}

\textbf{Mean Reciprocal Rank (MRR)} evaluates the ranking quality of a system by looking at the position of the first correct answer. Let $Q$ denote the set of queries and $\text{rank}_i$ the rank position of the first correct result for the $i$-th query. It is extensively used in \textbf{Link Prediction} and \textbf{Modality Retrieval} to assess the model's ability to prioritize the correct target, rewarding models that place the ground truth at the very top of the candidate list.
\begin{equation}
    \text{MRR} = \frac{1}{|Q|} \sum_{i=1}^{|Q|} \frac{1}{\text{rank}_i}.
\end{equation}

\textbf{Hits@K} measures the proportion of queries for which the correct target appears within the top-$K$ ranked results. This metric reflects recall performance at a fixed cutoff level. In the context of \textbf{Modality Retrieval} and \textbf{Link Prediction}, Hits@K reflects the system's effectiveness in recommending relevant candidates within a user-tolerable window (e.g., top-10), indicating the practical utility of the retrieval results.
\begin{equation}
    \text{Hits@K} = \frac{1}{|Q|} \sum_{i=1}^{|Q|} \mathbb{I}(\text{rank}_i \leq K).
\end{equation}

\textbf{Normalized Mutual Information (NMI)} quantifies the mutual dependence between the predicted cluster assignments $C$ and ground-truth labels $Y$, normalized by their entropies. Here, $I(Y; C)$ denotes mutual information, and $H(\cdot)$ denotes entropy. In \textbf{Node Clustering}, NMI is crucial for quantifying how much information the learned clusters share with ground-truth classes independent of label permutation, serving as a key indicator of disentanglement quality in unsupervised settings.
\begin{equation}
    \text{NMI}(Y, C) = \frac{2 \cdot I(Y; C)}{H(Y) + H(C)}.
\end{equation}

\textbf{Adjusted Rand Index (ARI)} evaluates the similarity between two data clusterings by considering all pairs of samples and counting pairs that are assigned in the same or different clusters in the predicted and predicted partitions. Let $n_{ij}$ be the number of samples in the intersection of ground-truth cluster $i$ and predicted cluster $j$, with marginal counts $a_i = \sum_j n_{ij}$ and $b_j = \sum_i n_{ij}$. Used alongside NMI for \textbf{Node Clustering}, ARI provides a stricter evaluation by adjusting for chance grouping, effectively penalizing random assignments and verifying the structural consistency of the discovered communities.
\begin{equation}
    \text{ARI} = \frac{\sum_{ij} \binom{n_{ij}}{2} - [\sum_i \binom{a_i}{2} \sum_j \binom{b_j}{2}] / \binom{n}{2}}{ \frac{1}{2} [\sum_i \binom{a_i}{2} + \sum_j \binom{b_j}{2}] - [\sum_i \binom{a_i}{2} \sum_j \binom{b_j}{2}] / \binom{n}{2} }.
\end{equation}

\textbf{CLIP-Score} evaluates cross-modal semantic consistency without the need for human annotations by leveraging pre-trained vision-language models. Let $E_I(\cdot)$ and $E_T(\cdot)$ denote the image and text encoders of a pre-trained CLIP model, respectively. For \textbf{Modality Alignment} and \textbf{Graph-to-Image Generation}, this metric acts as a semantic proxy, measuring whether the generated images or aligned features faithfully preserve the semantic content of the corresponding text or graph description.
\begin{equation}
    \text{CLIP-Score}(I, T) = \max\left(100 \cdot \cos(E_I(I), E_T(T)), 0\right).
\end{equation}

\textbf{BLEU-4} is a standard metric for evaluating generated text, focusing on the precision of $n$-grams. It calculates the geometric mean of modified $n$-gram precisions ($p_n$) up to length 4, multiplied by a brevity penalty (BP). In \textbf{Graph-to-Text} tasks, BLEU-4 evaluates the lexical accuracy and fluency of the generated descriptions, ensuring that the model produces phrases that precisely match the reference text.
\begin{equation}
    \text{BLEU-4} = \text{BP} \cdot \exp\left(\sum_{n=1}^4 w_n \log p_n\right).
\end{equation}

\textbf{ROUGE-L} evaluates the quality of text generation based on the Longest Common Subsequence (LCS) between the candidate and reference texts. Unlike n-gram precision metrics, ROUGE-L captures sentence-level structure and measures recall. For \textbf{Graph-to-Text} generation, particularly in summarization scenarios, it ensures that the generated text covers the comprehensive information content found in the ground truth.
\begin{equation}
    \text{ROUGE-L} = \frac{(1 + \beta^2) R_{lcs} P_{lcs}}{R_{lcs} + \beta^2 P_{lcs}}.
\end{equation}

\textbf{CIDEr} measures the consensus between a generated caption and a set of human references using TF-IDF weighting. It computes the cosine similarity of TF-IDF vectors ($g$) for $n$-grams, giving less weight to common, uninformative words. Specifically optimized for \textbf{Graph-to-Text} captioning, CIDEr correlates better with human judgment by emphasizing the distinctiveness and semantic importance of the generated terms.
\begin{equation}
    \text{CIDEr}_n(c, r) = \frac{1}{M} \sum_{i=1}^{M} \frac{g^n(c) \cdot g^n(r_i)}{\|g^n(c)\| \|g^n(r_i)\|}.
\end{equation}

\textbf{DINOv2-Score} evaluates visual similarity between a generated image $I_{\text{gen}}$ and a reference image $I_{\text{ref}}$, using feature embeddings extracted by a pre-trained DINOv2 encoder. Unlike CLIP-Score which focuses on semantics, DINOv2-Score assesses the visual fidelity, object layout, and structural consistency. In \textbf{Graph-to-Image} tasks, it ensures that the generated visual content maintains high perceptual quality and structural resemblance to real-world samples.
\begin{equation}
    \text{DINOv2-Score}(I_{\text{gen}}, I_{\text{ref}}) = \cos(\text{DINO}(I_{\text{gen}}), \text{DINO}(I_{\text{ref}})).
\end{equation}

\begin{table*}[t!]
\centering
\caption{Key model-specific hyperparameters for advanced multimodal graph learning models.}
\label{tab:hyper_mag}

\fontsize{8pt}{9pt}\selectfont
\renewcommand{\arraystretch}{1.15}

\makebox[0.7\textwidth][c]{%
\begin{tabular}{
p{0.20\textwidth}
p{0.36\textwidth}
p{0.26\textwidth}
}
\toprule
\textbf{Model} & \textbf{Key Hyperparameters} & \textbf{Search Space} \\
\midrule

\textbf{GraphMAE2}
& Masking Ratio $\rho$ \newline Re-masking Ratio $\rho_{re}$ \newline Re-masking Views $K$ \newline Loss Balance Weight $\lambda$
 & \{0.5\} \newline \{0.5\} \newline \{3\} \newline \{0.1, 1.0, 5.0, 10.0\} \\
\midrule

\textbf{DMGC}
& Homophilic Neighbors $k_l$ \newline Heterophilic Neighbors $k_h$ \newline Loss Weights $\lambda, \mu$
& \{10, 15, 20\} \newline \{2, 3, 4, 5, 6\} \newline \{0, $10^{-3}$, $10^{-1}$, 1, 10\} \\
\midrule

\textbf{MIG-GT}
 & Receptive Fields $K^{(M)}$ \newline Global Samples $C$ \newline Transformer Residual $\gamma$
 & \{1, 2, 3, 4\} \newline \{5, 10, 15, 20, 25\} \newline \{0.8, 0.9\} \\
\midrule

\textbf{DGF}
 & Filtering Coefficients $\alpha, \beta$ \newline Filtering Layers $T$ \newline Sampling Threshold $\theta$
 & \{0.01, 0.1, 1, 10, 100\} \newline \{5, 10, 15, 20, 25, 30\} \newline \{0.1, 0.2, \dots, 0.9\} \\
\midrule

\textbf{MMGCN}
 & GCN Layers $L$ \newline Dropout Rate $p$ \newline Loss Weights $\alpha, \beta, \gamma$
 & \{4\} \newline \{0.4\} \newline \{0.1, 0.5, 0.7\} \\
\midrule

\textbf{MGAT}
 & Propagation Depth $L$ \newline Gate Mechanism Type \newline Attention Dropout $\rho_{att}$
 & \{1, 2, 3\} \newline \{Inner-Prod, Concat, Bi-Interact\} \newline \{0.1, 0.2, \dots, 0.8\} \\
\midrule

\textbf{GSMN}
 & Scaling Factor $\lambda$ \newline GCN Layers $L$ \newline Margin $\gamma$
 & \{5, 10, 20\} \newline \{1, 2\} \newline \{0.2\} \\
\midrule

\textbf{MHGAT}
 & Network Layers $L$ \newline Embedding Dimension $d$ \newline Dropout Rate $p$
 & \{2, 4\} \newline \{32, 64, 128\} \newline \{0.6\} \\
\midrule

\textbf{MGNet}
& Neighbors $K$ \newline Output Dimension $D_{out}$ \newline MGNet Layers $L$
& \{2, 4, 6, 8, 10, 12\} \newline \{20, 40, 60, \dots, 120\} \newline \{1, 2, 3\} \\
\midrule

\textbf{LGMRec}
& Hyperedges $A$ \newline Fusion Weight $\alpha$ \newline Layers $K, H$
& \{1, 2, 4, \dots, 256\} \newline \{0.1, 0.2, \dots, 1.0\} \newline \{1, 2, 3, 4\} \\
\midrule

\textbf{UniGraph2}
 & Masking Ratio $\rho$ \newline Number of Experts $N_{E}$ \newline SPD Loss Weight $\lambda$
 & \{0.8\} \newline \{8\} \newline \{0.1\} \\
\midrule

\textbf{GraphGPT-O}
 & Visual Loss Weight $\lambda_{vm}$ \newline Text Loss Weight $\lambda_{lm}$ \newline Warmup Ratio $\alpha$
 & \{5\} \newline \{1\} \newline \{3e-3\} \\
\midrule

\textbf{MLaGA}
 & Projector Depth $L_{proj}$ \newline Aligner Neighbor Samples $K_{agg}$ \newline NC Demonstration Top-k $K_{demo}$
 & \{2\} \newline \{5\} \newline \{3\} \\
\midrule

\textbf{Graph4MM}
 & Diffusion Steps $K$ \newline Diffusion Factor $\alpha$ \newline MM-QFormer Blocks
 & \{0, 1, 2, 3, 4\} \newline \{0.1\} \newline \{1\} \\
\midrule

\textbf{NTSFormer}
 & Transformer Layers $L^{(tf)}$ \newline Routed Experts $M$ \newline Self-Teaching Weight $\lambda$
 & \{1, 2, 3, 4\} \newline \{2, 4, 6, 8\} \newline \{1.0\} \\
\midrule

\textbf{InstructG2I}
& Graph Guidance Scale $\lambda_G$ \newline Text Guidance Scale $\lambda_T$ \newline PPR Restart Probability $\alpha$
& \{0.8, 1.0\} \newline \{5.0, 7.5\} \newline \{0.15\} \\

\bottomrule
\end{tabular}
}
\end{table*}

\section{Experiment Environment}
\label{app:env}

The experiments are conducted on a workstation equipped with Intel Xeon Scalable processors and NVIDIA RTX 6000 Ada Generation GPUs (96GB VRAM), supported by 256GB of system RAM. The environment operates with CUDA 12.9. As for software versions, we utilize Python 3.10.18 and PyTorch 2.8.

\section{Hyperparameter Settings}
\label{app:hyperparameters}

In this section, we describe the hyperparameter configurations used in our experiments. To ensure fairness and reproducibility, we distinguish between task-level hyperparameters, which are shared across models within the same downstream task, and model-level hyperparameters, which are specific to individual architectures.

\subsection{Task-level Hyperparameters}
\label{app:task_hyperparameters}

For Graph-Based Tasks, we adhere to specific configurations to optimize topological learning. In Node Classification and Node Clustering, the learning rate is set to $5 \times 10^{-3}$ with a batch size of 512, and the weight decay is maintained at $1 \times 10^{-5}$. We sample 25 neighbors for aggregation to capture sufficient local structural information. While Node Classification requires 100 epochs for convergence, Node Clustering extends the training to 500 epochs to stabilize the self-supervised objectives. Conversely, for Link Prediction, the learning rate is adjusted to $1 \times 10^{-3}$ with a significantly larger batch size of 2048 to handle edge pairs efficiency, utilizing 15 neighbors per hop over 100 epochs.

For Modality-Based Tasks, configurations are categorized into alignment-based and generative processes. For alignment tasks, including Modality Matching, Retrieval, and Alignment, we utilize contrastive learning objectives with a temperature scaling factor $\tau$ of 0.07. These models are trained for 500 epochs with a learning rate of $1 \times 10^{-3}$ and a batch size of 256, utilizing early stopping with a patience ranging from 10 to 25 epochs to prevent overfitting. For generative tasks, settings are tailored to handle high-dimensional synthesis. In G2Text, following the experimental frameworks of MMGL~\cite{MMGL} and Graph4MM~\cite{Graph4MM}, we utilize OPT-125M and LLaMA-3.2-1B-Instruct backbones paired with a frozen CLIP-ViT-Large-Patch14 visual encoder and an XLM-RoBERTa text encoder. The models are trained for 20 epochs using a per-device batch size of 4 and a 200-step warmup. The global learning rate is set to $1 \times 10^{-3}$, with the projector learning rate scaled to $1 \times 10^{-2}$ and a weight decay of $1 \times 10^{-2}$. Visual features are mapped to 256 tokens for LLaMA, whereas they are adaptively pooled to 4 tokens for OPT to accommodate capacity constraints. We also provide details for different fine-tuning strategies. For Adapters, we employ IA3~\cite{ia3} with a boosted learning rate of $5 \times 10^{-3}$, targeting q\_proj, v\_proj, and fc2 modules for OPT, and k\_proj, v\_proj, and down\_proj for LLaMA. For LoRA, we apply low-rank adaptation with $r=64$ and $\alpha=128$ to the query and value matrices. In G2Image~\cite{instructg2i}, the learning rate is set to $1 \times 10^{-4}$ with a batch size of 16 for 20 epochs. Following the InstructG2I framework, this task adopts Semantic PPR-based sampling with neighbor numbers varying from 0 to 6, and sets the image resolution to 256 to condition the Stable Diffusion v1.5 backbone.

Additionally, for both scenarios, we standardize fundamental architectural parameters to ensure fairness. Unless otherwise specified for encoder-specific analyses, we employ clip-vit-large-patch14~\cite{clip} as the default feature encoder, unifying the node embedding dimensionality at 768 across all tasks. Optimization is conducted using the Adam or AdamW~\cite{adamw} optimizer. To evaluate the robustness of our results and mitigate the impact of initialization randomness, we eliminate the use of fixed random seeds for performance reporting. All experiments are repeated three times, and we report the mean results of the respective metrics for unbiased predictive performance.

\subsection{Model-level Hyperparameters}
\label{app:model_hyperparameters}

Model-level hyperparameters are tuned individually to align with the specific architectural designs of each method. For conventional GNNs, we generally configure the architecture with 3 layers and a hidden dimensionality of 256 to balance expressiveness and computational efficiency. Exceptions include deeper architectures like GCNII, which utilizes 8 layers, and parameter-efficient models like RevGAT and GATv2, which employ 2 layers with a hidden dimension of 128. Dropout rates are adjusted based on model sensitivity, ranging from 0.02 for GraphSAGE to 0.5 for models like RevGAT and GCNII. For attention-based mechanisms, the number of attention heads is selected from \{3, 4\}, with attention dropout typically set to 0.1. Additionally, method-specific hyperparameters are fixed to standard values, such as the filter size $K=3$ for ChebNet and smoothing parameters $\alpha=0.1, \theta=0.5$ for GCNII.

For advanced MAG learning models, we focus on tuning model-specific hyperparameters that are critical to each method’s design. Table~\ref{tab:hyper_mag} summarizes the key hyperparameters and their corresponding search spaces for the evaluated MAG models.

\section{Theoretical Complexity Analysis Details}
\label{app:efficiency}

To provide a comprehensive understanding of the efficiency profiles of MAG models, we present a detailed theoretical complexity analysis in Table~\ref{tab:complexity}. We adhere to standard notations where $|\mathcal{V}|$ and $|\mathcal{E}|$ represent the number of nodes and edges, respectively. $d$ denotes the hidden feature dimension, and $L$ indicates the number of GNN layers. For multimodal settings, $M$ represents the number of modalities. Regarding MLLM-based methods, $S$ denotes the sequence length of tokens, and $\Theta$ refers to the parameter size of the Large Language Model (LLM). Furthermore, specific hyperparameters such as the number of clusters or iterations are denoted by $K$, while $\Theta_{enc}$ represents the parameters of specific encoders.

Graph-enhanced models typically exhibit efficient scalability, primarily governed by the message-passing mechanism which scales linearly with respect to the number of edges $O(|\mathcal{E}|)$. Most methods in this category, such as GraphMAE2 and DGF, follow this pattern by leveraging sparse matrix operations, making them suitable for large-scale graphs. However, methods involving explicit structure learning or dense clustering, such as DMGC, introduce quadratic computational complexity $O(|\mathcal{V}|^2)$ due to the pairwise similarity calculations required to refine the graph topology. While this enables the model to capture latent structural dependencies, the resulting memory and computational overhead can become a bottleneck.

Multimodal-enhanced models generally incur higher computational costs proportional to the number of modalities $M$. These models often employ complex fusion mechanisms, such as multi-head attention in MGAT or MHGAT, which scale linearly with $M$ but may involve $O(d^2)$ terms due to attention matrix computations. While standard fusion operations maintain manageable complexity comparable to conventional GNNs, certain approaches like GSMN that explicitly model cross-modal interactions or perform dense metric learning can exhibit quadratic complexity $O(|\mathcal{V}|^2)$ or significant memory overhead. Consequently, the efficiency of this category is largely determined by the trade-off between the depth of cross-modal interaction and the efficiency of the fusion strategy.

MLLM-enhanced models represent the most computationally intensive category, dominated by the underlying Transformer architecture and the scale of the LLM. The complexity is largely determined by the self-attention mechanism, which scales quadratically with the sequence length $O(S^2)$, and the massive number of parameters $\Theta$ that impose substantial memory footprints. While some methods like MLaGA and UniGraph2 attempt to mitigate this by efficient alignment or adapter tuning to keep the complexity related to graph size linear, generative approaches like GraphGPT-O that deal with long context windows inevitably face significant training and inference overheads compared to traditional GNNs, necessitating high-end hardware resources for deployment.


\end{document}